\documentclass[]{bytedance_seed}

\usepackage[toc,page,header]{appendix}

\usepackage{minitoc}
\usepackage{natbib}
\usepackage{latexsym}

\usepackage{url}
\usepackage{amssymb}
\usepackage[utf8]{inputenc}
\usepackage{microtype}
\usepackage{booktabs}
\usepackage{pifont} 
\usepackage{multirow}
\usepackage{makecell}
\usepackage{paralist}
\usepackage{xspace}
\usepackage{color}
\usepackage{xcolor}
\usepackage{colortbl}
\usepackage{adjustbox}
\usepackage{hyperref} 
\usepackage[edges]{forest}
\usepackage{tikz} 
\usepackage{caption}
\usepackage{amsfonts}

\hypersetup{
    colorlinks,
    linkcolor={blue!80!black},
    citecolor={blue!80!black},
}
\tikzset{
    root/.style =             {align=center, text width=1cm, rounded corners=3pt, line width=0.3mm, fill=gray!10, draw=gray!80, font=\small},
    demographic/.style =         {align=center, text width=1.8cm, rounded corners=3pt, line width=0.3mm, fill=blue!10, draw=blue!80, font=\footnotesize},
    demographic_work/.style =    {align=center, text width=10cm, rounded corners=3pt, line width=0.3mm, fill=blue!10, draw=blue!0, font=\footnotesize},
    character/.style =         {align=center, text width=1.8cm, rounded corners=3pt, line width=0.3mm, fill=red!10, draw=red!80, font=\footnotesize},
    character_work/.style =    {align=center, text width=10cm, rounded corners=3pt, line width=0.3mm, fill=red!10, draw=red!0, font=\footnotesize},
    personalization/.style =           {align=center, text width=1.8cm, rounded corners=3pt, line width=0.3mm, fill=cyan!10, draw=cyan!80, font=\footnotesize},
    personalization_work/.style =      {align=center, text width=10cm, rounded corners=3pt, line width=0.3mm, fill=cyan!10, draw=cyan!0, font=\footnotesize},
    risk/.style =         {align=center, text width=1.8cm, rounded corners=3pt, line width=0.3mm, fill=orange!10, draw=orange!80, font=\footnotesize},
    risk_work/.style =    {align=center, text width=10cm, rounded corners=3pt, line width=0.3mm, fill=orange!10, draw=orange!0, font=\footnotesize},
}

\usepackage{CJK}
\title{UP: Unbounded Positive Asymmetric  Optimization for Breaking the Exploration-Stability Dilemma}

\author[1,2,*, \dagger]{Chongyu Fan}
\author[1]{Pengfei Liu}
\author[1]{Jingjia Huang}
\author[2]{Sijia Liu}
\author[1, \dagger]{Yi Lin}

\affiliation[1]{ByteDance Seed}
\affiliation[2]{Michigan State University}

\contribution[*]{Work done at ByteDance Seed}
\contribution[\dagger]{Corresponding authors}

\abstract{
Reinforcement learning (RL) has become the standard paradigm for enhancing the complex reasoning capabilities of large language models (LLMs). To achieve sample efficiency, modern RL frameworks rely on importance sampling (IS). However, these algorithms suffer from an exploration-stability dilemma. Pure IS often leads to catastrophic training instability, while standard clipping mechanisms used to mitigate this instability strictly constrain the policy update budget. By formalizing the concept of \textbf{Probability Capacity (Cap),} we reveal that conservative clipping structurally stifles exploration by prematurely truncating the update budget for correct but low-confidence reasoning paths. To break free from these constraints, we propose \textbf{Unbounded Positive Asymmetric Optimization (UP)}, a universal and plug-and-play objective. UP theoretically restructures the optimization process by anchoring the policy to its current state via the stop-gradient operator. This asymmetric design unleashes unclipped, stable gradients for positive advantages to maximize exploration, while maintaining standard clipping safeguards for negative advantages to prevent training instability. Furthermore, our formulation readily extends across different optimization granularities, including token-level (GRPO, DAPO) and sequence-level (GSPO) frameworks. Extensive experiments demonstrate that UP enhances exploration capacity and achieves superior reasoning accuracy across diverse RL algorithms (DAPO, GSPO, and GRPO), model architectures (Dense, MoE, and vision-language), and training modalities (language and multimodal), validating UP as a truly universal plug-and-play enhancement for RL-based training.
}

\date{\today}
\correspondence{Chongyu Fan at \email{fanchon2@msu.edu}, Yi Lin at \email{linyi.james@bytedance.com}}

\checkdata[Project Page]{\url{https://chongyu-fan.netlify.app/posts/up/}}

\begin{document}
\maketitle

%不需要目录就注释掉 注意目录不要和第一页放在一块 要有\newpage
%\newpage
%\tableofcontents
%\newpage

% \input{sections/outline}
% \clearpage
% \newpage

\section{Introduction}
\label{sec: intro}

With the rapid advancement of large language models (LLMs), their ability to solve complex, multi-step mathematical and logical reasoning tasks has become increasingly prominent~\citep{guo2025deepseek,yang2025qwen3,chen2025minimax,seed2026seed2}. However, navigating the vast and intricate search space of reasoning trajectories using supervised fine-tuning alone is often infeasible~\citep{zelikman2022star,lightman2023let}. To address this challenge, reinforcement learning (RL) has emerged as the standard paradigm~\citep{ouyang2022training,bai2022constitutional}. While foundational on-policy methods like REINFORCE~\citep{williams1992simple} offer mathematically stable optimization, they suffer from severe sample inefficiency. Consequently, modern RL frameworks such as Group Relative Policy Optimization (GRPO)~\citep{shao2024deepseekmath} and Dynamic sAmpling Policy Optimization (DAPO)~\citep{yu2025dapo} have transitioned to multi-step optimization, which aims to improve sample efficiency by utilizing \textit{importance sampling (IS)} to estimate target distributions from historical policies ($\pi_{\text{old}}$). More recently, a growing body of work has extended the GRPO/DAPO paradigm along diverse directions, including sequence-level optimization via GSPO~\citep{zheng2025group}, critic-free global normalization via REINFORCE++~\citep{hu2025reinforce++}, and asymmetric importance sampling designs such as TOPR~\citep{roux2025tapered} and ASPO~\citep{wang2025aspo}.

Despite the widespread adoption of IS-based RL algorithms, recent theoretical and empirical observations have identified a critical issue: \textit{IS-based RL inherently suffers from the exploration-stability dilemma}~\citep{liu2025understanding,zheng2025prosperity,yue2025does,ma2025stabilizing,fan2026rethinking}. Specifically, pure unclipped IS is highly susceptible to pathological gradient explosion. When evaluating rare, long-tail reasoning paths, the IS ratio can explode, leading to catastrophic \textit{training instability}~\citep{schulman2017proximal,schulman2015trust}. To mitigate this instability, standard algorithms heavily rely on a \textit{clipping} mechanism, which forcibly bounds the policy update within a predefined trust region to preserve original model representations~\citep{schulman2017proximal,li2026bandpo}. 

The clipping mechanism in standard RL algorithms inherently restricts the allowable changes in token probabilities. Although some recent studies have recognized this limitation~\citep{su2025klear,gao2025soft,zhang2025design,fan2026rethinking}, they have not provided a fundamental resolution. To formally quantify this constraint, we define the \textbf{Probability Capacity (Cap)} as the maximum allowable increase in $\pi_\theta$ for tokens with positive advantages, or the maximum allowable decrease for tokens with negative advantages, before the optimization gradient is truncated. For example, DAPO, one of the state-of-the-art algorithms, attempts to mitigate exploration stagnation by utilizing an elevated upper clip bound ($\epsilon_{\text{high}}$) for positive advantages. However, as we formally show in Sec.\,\ref{sec:preliminary}, DAPO still remains vulnerable to stifled exploration. By analyzing the Cap, we reveal that adjusting $\epsilon_{\text{high}}$ only alleviates the issue without fundamentally resolving it. Because the clipping mechanism strictly restricts the Cap for correct but low-confidence logic paths to be linearly proportional to the historical policy ($\pi_{\text{old}}$), these tokens still receive a severely constrained update budget. Once the policy marginally improves, the gradient is prematurely truncated to zero. This highlights the need to develop a new mathematical foundation to break free from conservative clipping constraints and fully unleash the model's exploration~\citep{yue2025does,li2026back,cao2025efficient}.

Motivated by this observation, we ask:
\begin{tcolorbox}[before skip=2mm, after skip=0.0cm, boxsep=0.0cm, middle=0.0cm, top=0.05cm, bottom=0.05cm, boxrule=0.6pt]
\begin{center}
     \textit{\textbf{(Q)} How can we maximize a model's exploration while preventing training instability in RL?}
\end{center}
\end{tcolorbox} 
\vspace*{2mm}

Drawing inspiration from the stable, unclipped gradients of REINFORCE, we address \textit{(Q)} through the lens of \textbf{Unbounded Positive Asymmetric Optimization (UP)}. Specifically, the treatment of positive advantages focuses on maximizing exploration capacity, coupled with the treatment of negative advantages that acts as a structural safeguard against training instability~\citep{zhu2025surprising,wang2025aspo,roux2025tapered}. We theoretically restructure the importance sampling mechanism by replacing the IS ratio with a self-anchored ratio using the \textit{stop-gradient operator (sg)}~\citep{zhang2025design}. 
We show that this asymmetric formulation naturally aligns with an unclipped, mathematically stable gradient for correct rollouts. UP, by explicitly bypassing the traditional IS anchor, opens a critical yet underexplored direction for enhancing RL reasoning capabilities without sacrificing optimization stability. 

We summarize our \textbf{contributions} below. 

$\bullet$ We formalize the concept of \textbf{Probability Capacity (Cap)} to expose a fundamental dilemma in current RL algorithms: overly conservative clipping constraints structurally stifle exploration for reasoning, whereas overly aggressive updates inevitably lead to training instability.

$\bullet$ We theoretically restructure the optimization process by anchoring the policy to its current state via the stop-gradient operator. Coupled with an in-depth exploration of asymmetric optimization design, we propose \textbf{Unbounded Positive Asymmetric Optimization (UP)}, a universal, plug-and-play objective that readily extends across different optimization granularities (e.g., token-level (GRPO, DAPO) and sequence-level (GSPO)).

$\bullet$ We conduct extensive experiments to demonstrate the critical role of the UP formulation in resolving the exploration-stability dilemma. Empirically, UP consistently enhances exploration capacity and achieves superior accuracy across diverse RL algorithms (DAPO, GSPO, and GRPO), model architectures (Dense, MoE, and vision-language), and training modalities (language and multimodal), while preventing training instability across all settings. In addition, against eleven strong RL baselines, including GRPO, Dr.\,GRPO, CISPO, DPPO, GMPO, GSPO, SAPO, REINFORCE++, RLOO, W-REINFORCE, and ASPO, UP-GRPO attains the best average Pass@1 accuracy across five challenging reasoning benchmarks (AIME24, AMC23, MATH500, Minerva, and OlympiadBench).
\section{Related Work}

\noindent \textbf{Reinforcement Learning for LLM Reasoning.} 
While REINFORCE \cite{williams1992simple} established the core policy gradient framework, PPO \cite{schulman2017proximal} became the standard RLHF/RLAIF paradigm \cite{ouyang2022training, bai2022constitutional} by improving sample efficiency. However, PPO's auxiliary critic model limits scalability for long-context reasoning. To alleviate this, GRPO \cite{shao2024deepseekmath} omits the critic via group-level relative scoring, driving breakthroughs in models like DeepSeek-R1 \cite{guo2025deepseek} and Qwen2.5-Math \cite{yang2024qwen2}. Building on GRPO, DAPO \cite{yu2025dapo} introduces decoupled clipping boundaries, while GSPO \cite{zheng2025group} and related works \cite{mao2025clip, liu2026length} transition to sequence-level importance sampling (IS) to reduce variance. Despite these advancements, current paradigms still inherit restrictive symmetric clipping mechanisms \cite{liu2025understanding}, which we argue structurally stifle exploration.

\noindent \textbf{Importance Sampling and Asymmetric Optimization.}
To prevent training instability caused by exploding IS ratios, TRPO \cite{schulman2015trust} and PPO introduced heuristic clipping. While recent efforts (e.g., BandPO \cite{li2026bandpo}, SAPO \cite{gao2025soft}, GMPO \cite{zhao2025geometric}, and M2PO \cite{zheng2025prosperity}) refine these bounds, they all assume the historical policy ($\pi_{\text{old}}$) must remain in the denominator. This assumption bottlenecks exploration, the most critical challenge in reasoning where golden trajectories are sparse \cite{zelikman2022star, snell2024scaling, fan2026cyclicreflex}, rendering the discovery of these rare paths futile if their gradients are immediately clipped during optimization. Consequently, asymmetric optimization has gained traction to address these conflicting signals (e.g., ASPO \cite{wang2025aspo} and W-REINFORCE \cite{zhu2025surprising}). Our UP distinctly advances this paradigm: by introducing a stop-gradient self-anchor ($\text{sg}(\pi_\theta)$), we theoretically restructure the IS ratio to bypass the trust region bottleneck. By combining this Unbounded Positive (UP) formulation for correct rollouts with DAPO's rigorous negative constraints, our method aggressively internalizes long-tail reasoning capabilities without triggering IS instability.
\section{The Exploration-Stability Dilemma in Importance Sampling and Clipping}
\label{sec:preliminary}

In this section, we analyze the mathematical framework of RL paradigms like REINFORCE and GRPO, exposing the structural dilemma between importance sampling-induced training instability and clipping-induced exploration stagnation in LLM reasoning.

\subsection{Definitions and Formulations: REINFORCE, GRPO, DAPO, and GSPO}

\noindent \textbf{REINFORCE.} The evolution of RL algorithms is fundamentally driven by the pursuit of sample efficiency and optimization stability. As a foundational approach, the vanilla REINFORCE algorithm optimizes the policy strictly on-policy \cite{williams1992simple}. Given a query $q$ and a generated response $o_{i,t}$ at step $t$, the REINFORCE objective is defined as maximizing the expected advantage-weighted log-probability:
\begin{equation}
\mathcal{J}_{\text{REINFORCE}}(\theta) = \mathbb{E}_{q \sim \mathcal{Q}, o \sim \pi_\theta} \left[ \sum_{t=1}^{|o|} \hat{A} \log \pi_\theta(o_{i,t}|q, o_{i,<t}) \right] \label{eq:reinforce_obj}
\end{equation}
Taking the derivative of this objective yields a mathematically stable gradient:
\begin{equation}
\nabla_\theta \mathcal{J}_{\text{REINFORCE}}(\theta) = \mathbb{E}_{q \sim \mathcal{Q}, o \sim \pi_\theta} \left[ \sum_{t=1}^{|o|} \hat{A} \nabla_\theta \log \pi_\theta(o_{i,t}|q, o_{i,<t}) \right] \label{eq:reinforce_grad}
\end{equation}

\noindent \textbf{GRPO and DAPO.} While mathematically stable, REINFORCE suffers from severe sample inefficiency because trajectories must be discarded after a single gradient step. To enable multi-step optimization over the same sampled data, modern frameworks like Group Relative Policy Optimization (GRPO) introduce \textbf{Importance Sampling (IS)} \cite{shao2024deepseekmath}. Mathematically, IS is a statistical technique used to estimate the properties of a target distribution (the current policy $\pi_\theta$) using samples drawn from a distinct proposal distribution (the historical policy $\pi_{\text{old}}$). This is achieved via the importance sampling ratio:
\begin{equation}
r_{i,t}(\theta) = \frac{\pi_\theta(o_{i,t}|q, o_{i,<t})}{\pi_{\text{old}}(o_{i,t}|q, o_{i,<t})}
\end{equation}

Despite its sample efficiency, pure unclipped IS suffers from severe optimization instability, as we formally demonstrate in \textbf{Sec.\,\ref{subsec:dilemma_1}}. To mitigate this instability, GRPO introduces a \textbf{clipping} operation. This heuristic regularization technique is designed to restrict the policy update step size by forcibly bounding the importance sampling ratio within a predefined trust region $[1-\epsilon, 1+\epsilon]$. The full GRPO objective is formulated as:
\begin{equation}
\mathcal{J}_{\text{GRPO}}(\theta) = \mathbb{E}_{q \sim \mathcal{Q}, \{o_i\}_{i=1}^G \sim \pi_{\text{old}}} \left[ \frac{1}{G} \sum_{i=1}^G \frac{1}{|o_i|} \sum_{t=1}^{|o_i|} \left\{ \min \left[ r_{i,t}(\theta) \hat{A}_i, \text{clip}(r_{i,t}(\theta), 1-\epsilon, 1+\epsilon) \hat{A}_i \right] - \beta \mathbb{D}_{\text{KL}}(\pi_\theta \| \pi_{\text{ref}}) \right\} \right] \label{eq:grpo_obj}
\end{equation}
where $\hat{A}_i$ is the group-normalized advantage. Building upon this foundation, state-of-the-art methods like Dynamic sAmpling Policy Optimization (DAPO) introduce decoupled clip bounds ($\epsilon_{\text{low}}, \epsilon_{\text{high}}$) while entirely omitting the KL penalty ($\beta = 0$) \cite{yu2025dapo}. Consequently, these methods rely exclusively on clipping to balance optimization stability and enhance exploration capacity, exposing a deep mathematical vulnerability.

\noindent \textbf{GSPO.} To address the inherent mismatch between sequence-level rewards and the token-level optimization utilized in GRPO and DAPO, Group Sequence Policy Optimization (GSPO) was recently proposed. GSPO transitions to a sequence-level importance sampling framework \cite{zheng2025group}. Unlike GRPO, which computes the importance ratio $r_{i,t}(\theta)$ and applies clipping for each individual token, GSPO defines the importance ratio based on the length-normalized likelihood of the entire generated sequence. To prevent gradient variance explosion associated with long reasoning trajectories, GSPO calculates the geometric mean of the token-level ratios. Let $\pi_\theta(o_i|q) = \prod_{t=1}^{|o_i|} \pi_\theta(o_{i,t}|q, o_{i,<t})$ denote the likelihood of the generated response. The length-normalized sequence-level importance ratio $s_i(\theta)$ is defined as:
\begin{equation}
s_i(\theta) = \left( \frac{\pi_\theta(o_i|q)}{\pi_{\text{old}}(o_i|q)} \right)^{\frac{1}{|o_i|}} = \exp \left( \frac{1}{|o_i|} \sum_{t=1}^{|o_i|} \left( \log \pi_\theta(o_{i,t}|q, o_{i,<t}) - \log \pi_{\text{old}}(o_{i,t}|q, o_{i,<t}) \right) \right)
\end{equation}
The GSPO objective applies clipping, rewarding, and optimization at the sequence level by directly constraining this aggregated ratio $s_i(\theta)$:
\begin{equation}
\mathcal{J}_{\text{GSPO}}(\theta) = \mathbb{E}_{q \sim \mathcal{Q}, \{o_i\}_{i=1}^G \sim \pi_{\text{old}}} \left[ \frac{1}{G} \sum_{i=1}^G \min \left( s_i(\theta) \hat{A}_i, \text{clip}(s_i(\theta), 1-\epsilon, 1+\epsilon) \hat{A}_i \right) \right] \label{eq:gspo_obj}
\end{equation}

\subsection{Aggressive Updates Dilemma 1: Importance Sampling Induces Training Instability}
\label{subsec:dilemma_1}

To understand why clipping is considered strictly necessary in the aforementioned methods, we must first formalize the fatal bottleneck introduced by pure IS. The unclipped IS objective seeks to maximize the expected advantage weighted by the importance sampling ratio $r_{i,t}(\theta)$:
\begin{equation}
\mathcal{J}_{\text{IS}}(\theta) = \mathbb{E}_{q \sim \mathcal{Q}, o \sim \pi_{\text{old}}} \left[ \hat{A} r_{i,t}(\theta) \right] \label{eq:is_obj_pure}
\end{equation}

Applying the log-derivative trick, the gradient of this unclipped objective becomes:
\begin{equation}
\nabla_\theta \mathcal{J}_{\text{IS}}(\theta) = \mathbb{E}_{q \sim \mathcal{Q}, o \sim \pi_{\text{old}}} \left[ \hat{A} r_{i,t}(\theta) \nabla_\theta \log \pi_\theta(o_{i,t}|q, o_{i,<t}) \right] \label{eq:is_grad}
\end{equation}

Directly comparing \textbf{Eq.\,\ref{eq:is_grad}} to the REINFORCE gradient in \textbf{Eq.\,\ref{eq:reinforce_grad}}, the mathematical flaw is evident: the IS gradient is explicitly scaled by $r_{i,t}(\theta)$. For rare, high-reward reasoning paths where the behavior probability $\pi_{\text{old}}(o_{i,t}|q, o_{i,<t})$ is infinitesimally small, $r_{i,t}(\theta)$ explodes as the active policy marginally improves. This disproportionate scaling injects pathological gradients, irrevocably destroying stable representations and inevitably leading to severe \textbf{training instability}.

\begin{figure*}[htbp]
\centering
\begin{tabular}{cccc}
% 第一行：放置 Legend，居中并跨越 4 列
% \hspace{-3mm}
\multicolumn{4}{c}{
    \includegraphics[width=0.96\textwidth]{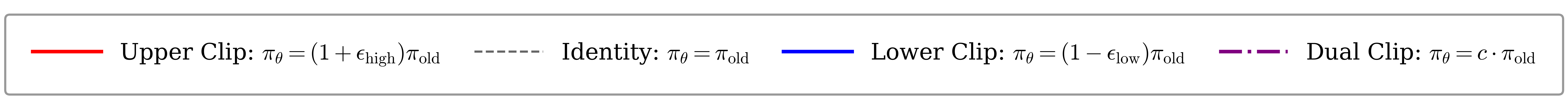}
} \\
\vspace{-4mm} \\ % 微调 Legend 与下方图片的间距

% 第二行：a, b, c 图片和 Colorbar
\hspace*{-9mm}
\includegraphics[width=0.32\textwidth]{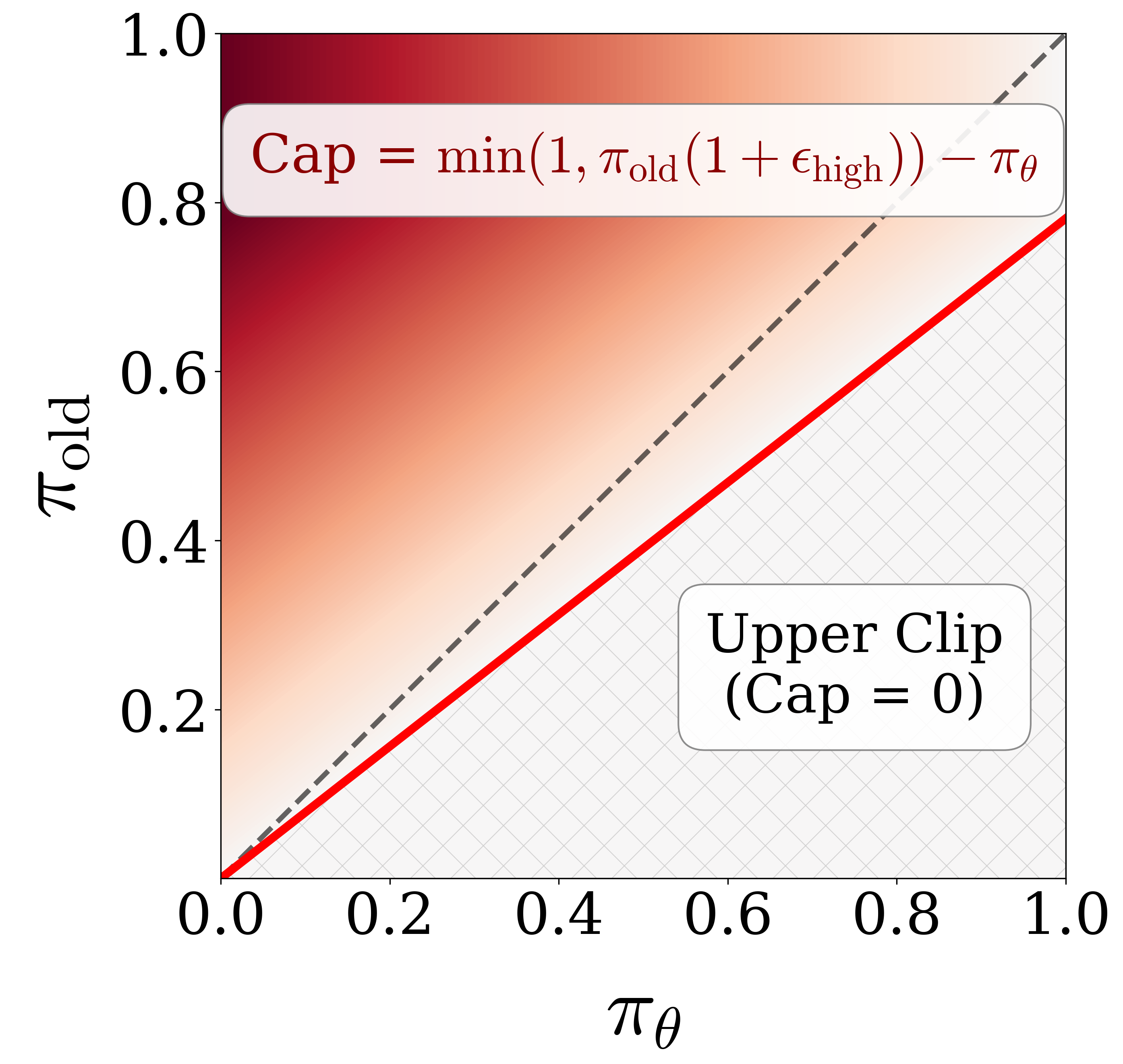}
&
\hspace*{-6mm}
\includegraphics[width=0.32\textwidth]{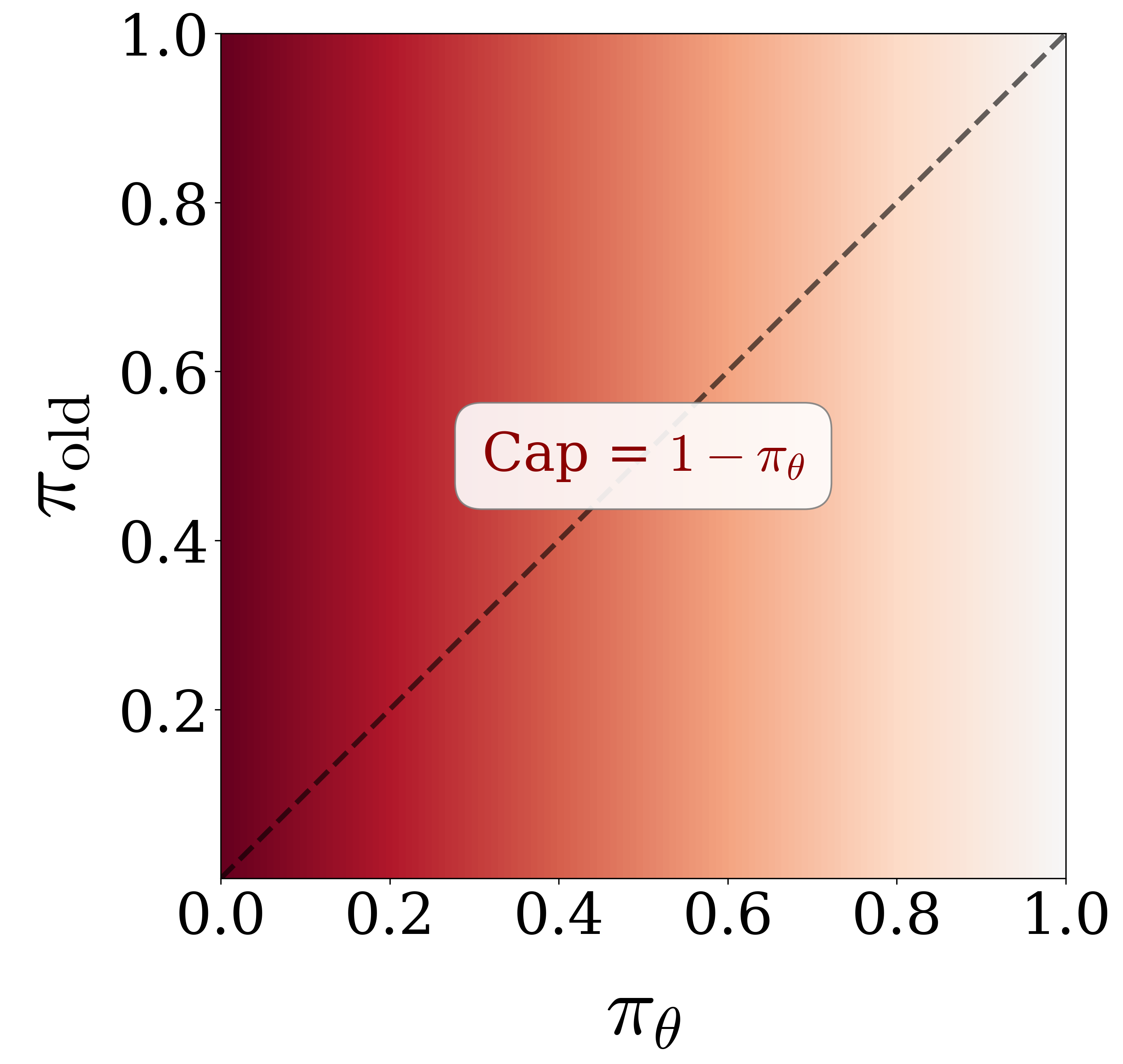}
&
\hspace*{-6mm}
\includegraphics[width=0.32\textwidth]{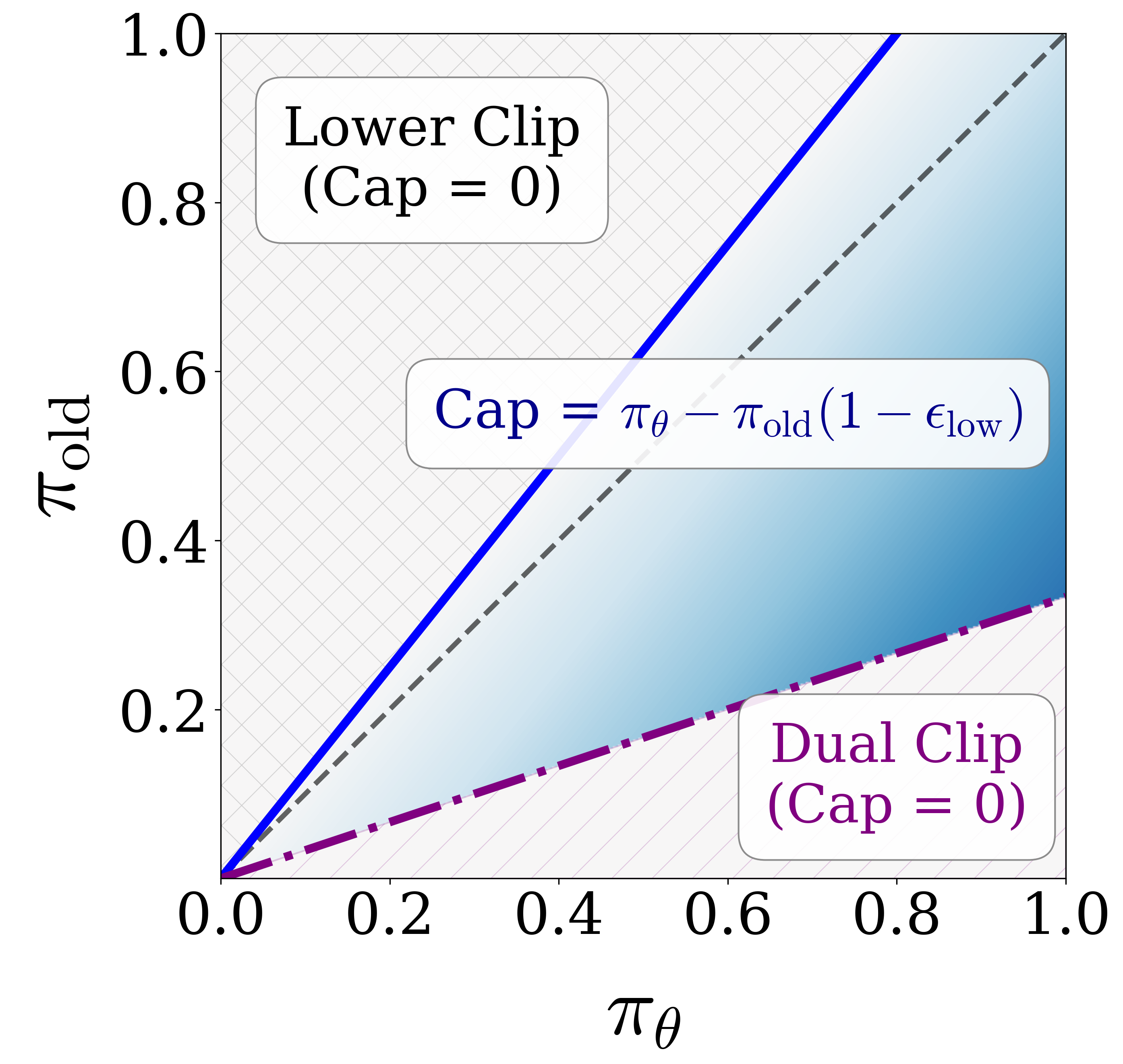}
&
\hspace*{-6mm}
\raisebox{7mm}{%
    \includegraphics[width=0.10\textwidth]{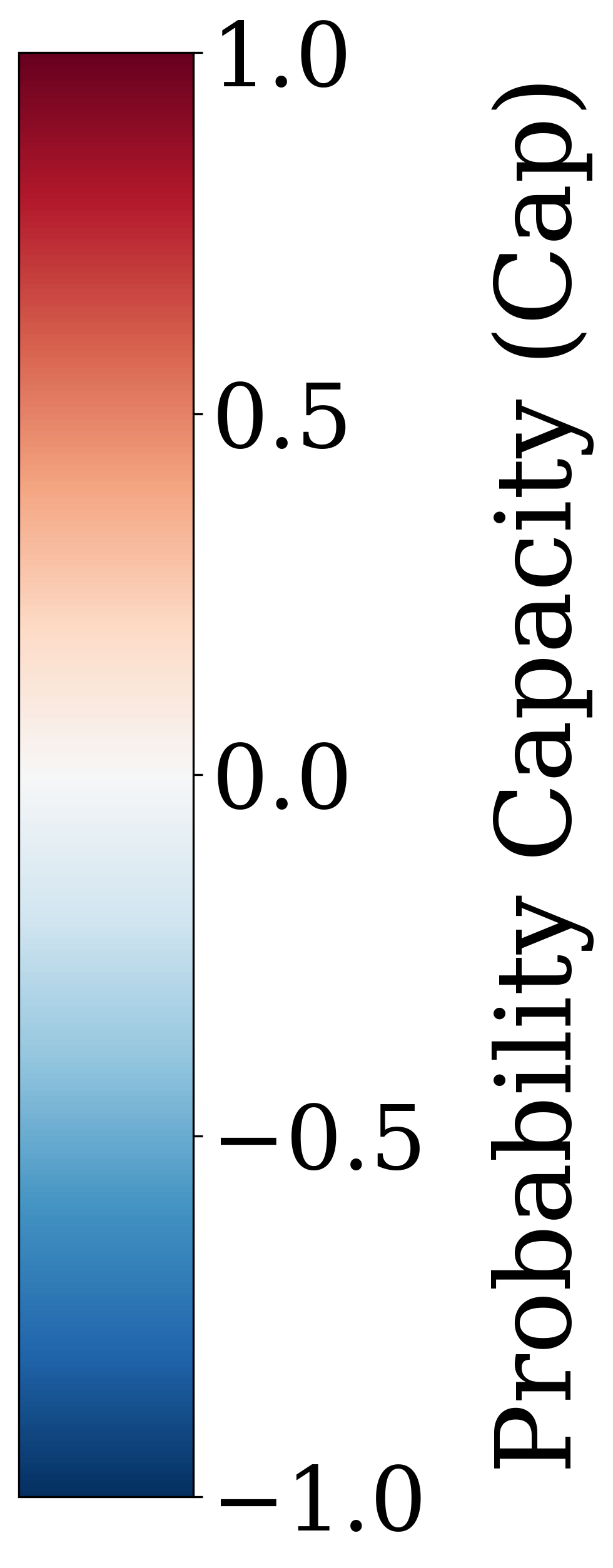}%
}
\\

% 第三行：标注 (a), (b), (c)
% \hspace*{-3mm}
\small{(a) DAPO ($\hat{A} > 0$)}
&
% \hspace*{-6mm}
\small{(b) UP-DAPO ($\hat{A} > 0$)}
&
% \hspace*{-6mm}
\small{(c) DAPO/UP-DAPO ($\hat{A} \leq 0$)}
&
% Colorbar 下方通常不标字母，或者留空
\\
\end{tabular}

\vspace*{-2mm}
\caption{\small{Visualization of the probability capacity (Cap) for DAPO and UP-DAPO. Cap is defined as the maximum allowable increase in $\pi_{\theta}$ for tokens with positive advantages ($\hat{A} > 0$), or the maximum allowable decrease for tokens with non-positive advantages ($\hat{A} \leq 0$). 
(a) DAPO under $\hat{A} > 0$, where $\epsilon_{\text{high}}$ constrains the update, resulting in $\text{Cap} = 0$ within the Upper Clip region. 
(b) UP-DAPO under $\hat{A} > 0$, where the absence of clipping allows all tokens to maintain a corresponding Cap. 
(c) DAPO and UP-DAPO under $\hat{A} \leq 0$, where $\epsilon_{\text{low}}$ and $c$ trigger Lower Clip and Dual Clip respectively, leading to $\text{Cap} = 0$ in those regions.}}
\label{fig:prob_capacity_comparison}
\end{figure*}

\subsection{Conservative Constraints Dilemma 2: Clipping Stifles Exploration}

To circumvent gradient explosion, standard algorithms like GRPO and DAPO depend heavily on clipping. Specifically, for positive advantages ($\hat{A}_i > 0$) where exploration is most critical for discovering novel and correct reasoning paths, the clipped objective $\mathcal{J}_{\text{IS+CLIP}}$ becomes:
\begin{equation}
\mathcal{J}_{\text{IS+CLIP}}(\theta) = \min \left[ r_{i,t}(\theta), 1 + \epsilon \right] \hat{A}_i \label{eq:clip_obj_pos}
\end{equation}

The gradient of this objective with respect to $\theta$ is:
\begin{equation}
\nabla_\theta \mathcal{J}_{\text{IS+CLIP}}(\theta) = \begin{cases} 
\hat{A}_i r_{i,t}(\theta) \nabla_\theta \log \pi_\theta(o_{i,t}|q, o_{i,<t}) & \text{if } r_{i,t}(\theta) \le 1 + \epsilon \\ 
0 & \text{if } r_{i,t}(\theta) > 1 + \epsilon 
\end{cases} \label{eq:clip_grad}
\end{equation}

To formally analyze the consequences of this gradient truncation (\textbf{Eq.\,\ref{eq:clip_grad}}), we define the \textbf{Probability Capacity (Cap)} as the maximum allowable increase (when $\hat{A} > 0$) or decrease (when $\hat{A} \le 0$) in $\pi_\theta$ in the absolute probability space before the update is blocked. This capacity represents the effective ``budget'' for policy exploration relative to the reference policy $\pi_{\text{old}}$.

As illustrated in \textbf{Fig.\,\ref{fig:prob_capacity_comparison}(a)}, for tokens with positive advantages ($\hat{A}_{i,t} > 0$) in DAPO, the gradient vanishes entirely the exact moment the probability ratio $r_{i,t}(\theta)$ exceeds the trust region boundary $1+\epsilon_{\text{high}}$. Mapping this constraint to the absolute probability space, the policy is strictly upper-bounded by $\pi_\theta \le (1+\epsilon_{\text{high}}) \pi_{\text{old}}$. Considering that predicted probabilities cannot exceed 1, the maximum achievable probability is $\min\big(1, (1+\epsilon_{\text{high}}) \pi_{\text{old}}\big)$. 

Consequently, the remaining Probability Capacity for a positive update can be explicitly formulated as a piecewise function:
\begin{equation}
\text{Cap}(\pi_\theta, \pi_{\text{old}}) = \begin{cases} 
\min\big(1, (1+\epsilon_{\text{high}})\pi_{\text{old}}\big) - \pi_\theta & \text{if } \pi_\theta < (1+\epsilon_{\text{high}})\pi_{\text{old}} \\ 
0 & \text{if } \pi_\theta \ge (1+\epsilon_{\text{high}})\pi_{\text{old}} 
\end{cases} \label{eq:capacity_piecewise}
\end{equation}

This piecewise formulation reveals a severe structural bottleneck: the update budget is linearly dependent on $\pi_{\text{old}}$. For low-likelihood reasoning tokens (e.g., $\pi_{\text{old}} = 0.01$), a standard threshold of $\epsilon_{\text{high}} = 0.28$ restricts the maximum absolute probability increase to a mere $0.0028$. Once $\pi_\theta$ reaches $0.0128$, the capacity drops to exactly zero (entering the ``Upper Clip'' region shown in Fig.\,\ref{fig:prob_capacity_comparison}(a)), and the gradient is nullified. Even if this specific action yields an exceptionally high advantage, its effective update is prematurely blocked. Ultimately, this mechanism structurally \textbf{stifles exploration}, preventing the algorithm from reinforcing promising long-tail trajectories and leaving optimal reasoning paths undiscovered.

\section{Breaking the Dilemma: Unbounded Positive Asymmetric Optimization}
\label{sec:method}

This section details how we dismantle the exploration-stability dilemma through two core mathematical innovations: replacing the historical policy $\pi_{\text{old}}$ with the current policy $\pi_\theta$ via the stop-gradient operator, and designing a dynamically routed asymmetric optimization framework. Together, these innovations constitute our \textbf{Unbounded Positive Asymmetric Optimization (UP)} methodology.

\subsection{Unbounded Formulation for Positive Advantages}

To eliminate the pathological gradients injected by $\pi_{\text{old}}$ while retaining the multi-step optimization over the same sampled data, we introduce \textbf{Unbounded Positive Asymmetric Optimization (UP)}. Crucially, the unbounded mechanism within UP is specifically designed to be applied exclusively to correct rollouts ($\hat{A} > 0$). For these correct rollouts, we propose replacing the standard importance sampling ratio $r_{i,t}(\theta)$ with a self-anchored modified ratio $\tilde{r}_{i,t}(\theta)$. This is achieved by utilizing the \textbf{stop-gradient operator ($\text{sg}$)} to explicitly formulate the unbounded positive component of our framework. This component is denoted as $\mathcal{J}_{\text{UP}}^+(\theta)$, where the superscript ``$+$'' indicates its restriction to correct rollouts. The objective is formulated as:
\begin{equation}
\mathcal{J}_{\text{UP}}^+(\theta) = \mathbb{E}_{q \sim \mathcal{Q}, o \sim \pi_{\text{old}}} \left[ \sum_{t=1}^{|o|} \hat{A} \tilde{r}_{i,t}(\theta) \right] = \mathbb{E}_{q \sim \mathcal{Q}, o \sim \pi_{\text{old}}} \left[ \sum_{t=1}^{|o|} \hat{A} \frac{\pi_\theta(o_{i,t}|q, o_{i,<t})}{\text{sg}(\pi_\theta(o_{i,t}|q, o_{i,<t}))} \right], \quad \text{for } \hat{A} > 0 \label{eq:up_obj}
\end{equation}
where $\tilde{r}_{i,t}(\theta) = \frac{\pi_\theta(o_{i,t}|q, o_{i,<t})}{\text{sg}(\pi_\theta(o_{i,t}|q, o_{i,<t}))}$.

During backpropagation, the term $\text{sg}(\pi_\theta(o_{i,t}|q, o_{i,<t}))$ functions strictly as a constant scalar. Because the denominator $\text{sg}(\pi_\theta)$ is equal to $\pi_\theta$ in value, taking the derivative of this objective allows us to seamlessly compute the gradient and apply the log-derivative trick ($\nabla x / x = \nabla \log x$) in a single continuous derivation:
\begin{equation}
\begin{aligned}
\nabla_\theta \mathcal{J}_{\text{UP}}^+(\theta) &= \mathbb{E}_{q \sim \mathcal{Q}, o \sim \pi_{\text{old}}} \left[ \sum_{t=1}^{|o|} \hat{A} \frac{1}{\text{sg}(\pi_\theta(o_{i,t}|q, o_{i,<t}))} \nabla_\theta \pi_\theta(o_{i,t}|q, o_{i,<t}) \right] \\
&= \mathbb{E}_{q \sim \mathcal{Q}, o \sim \pi_{\text{old}}} \left[ \sum_{t=1}^{|o|} \hat{A} \frac{1}{\pi_\theta(o_{i,t}|q, o_{i,<t})} \nabla_\theta \pi_\theta(o_{i,t}|q, o_{i,<t}) \right] \\
&= \mathbb{E}_{q \sim \mathcal{Q}, o \sim \pi_{\text{old}}} \left[ \sum_{t=1}^{|o|} \hat{A} \nabla_\theta \log \pi_\theta(o_{i,t}|q, o_{i,<t}) \right]
\end{aligned} \label{eq:up_grad_merged}
\end{equation}

This substitution yields a profound theoretical conclusion: optimizing this stable, unclipped ratio is mathematically equivalent to maximizing the REINFORCE objective established in \textbf{Eq.\,\ref{eq:reinforce_obj}}. By explicitly anchoring the policy to $\pi_\theta$ rather than $\pi_{\text{old}}$, this formulation completely eradicates the root cause of IS-induced instability. It safely enables unbounded reinforcement for golden reasoning trajectories without triggering the gradient explosion analyzed in \textbf{Sec.\,\ref{subsec:dilemma_1}}. As shown in \textbf{Fig.\,\ref{fig:prob_capacity_comparison}(b)}, the Cap under this formulation becomes fundamentally unconstrained by the historical policy. Defined simply by the maximum allowable increase ($\text{Cap} = 1 - \pi_\theta$), this ensures all tokens maintain a full update budget without clipping.

Building upon this unbounded formulation, we complete the UP framework. We intentionally engineer this asymmetric design to address the divergent mathematical dynamics between correct and wrong rollouts. For correct rollouts, our primary objective is to maximize the exploration capacity and substantially amplify the reinforcement signal for rare, low-confidence tokens. Therefore, we directly apply the unbounded positive component ($\mathcal{J}_{\text{UP}}^+$) to unleash unconstrained reinforcement. Conversely, for wrong rollouts, the advantage is inherently negative, which reverses the direction of the gradient. Applying an unbounded update performs aggressive gradient ascent and destroys the original representation. Consequently, this unbounded mechanism must be strictly prohibited in the negative regime.

\subsection{UP-GxPO: Universal Asymmetric Integration}

Crucially, the UP framework serves as a universal plugin that can be seamlessly integrated with any Group-based Policy Optimization (GxPO) algorithm. Recognizing these opposing requirements, we apply an asymmetric modification to standard policy objectives by dynamically routing the gradient computation based on the correctness of the rollouts.

In this work, we first instantiate our primary token-level method as \textbf{UP-DAPO} by combining the unbounded formulation for positive advantages with the DAPO baseline for negative advantages:
\begin{equation}
\begin{aligned}
\mathcal{J}_{\text{UP-DAPO}}(\theta) &= \mathbb{E}_{q \sim \mathcal{Q}, \{o_i\}_{i=1}^G \sim \pi_{\text{old}}} \Bigg[ \frac{1}{\sum_{i=1}^G |o_i|} \sum_{i=1}^G \sum_{t=1}^{|o_i|} \\
&\quad \begin{cases} 
    \hat{A}_{i,t} \log \pi_\theta(o_{i,t}|q, o_{i,<t}) & \text{if } \hat{A}_{i,t} > 0 \\
    \min \left( r_{i,t}(\theta) \hat{A}_{i,t}, \text{clip}(r_{i,t}(\theta), 1 - \epsilon_{\text{low}}, 1 + \epsilon_{\text{high}}) \hat{A}_{i,t} \right) & \text{if } \hat{A}_{i,t} \le 0
\end{cases} \Bigg],
\end{aligned}
\label{eq:up_dapo_obj}
\end{equation}
where the objective for wrong rollouts utilizes the standard decoupled clipping mechanism from DAPO to penalize incorrect actions.

This asymmetric design explicitly resolves the central dilemma analyzed in the previous section. For correct rollouts ($\hat{A} > 0$), we completely discard the clipping mechanism, actively overcoming the conservative constraints and the capacity mismatch identified previously. This deliberate unboundedness maximizes the exploration for rare, low-confidence tokens, allowing the model to aggressively reinforce successful long-tail reasoning paths. Conversely, for wrong rollouts ($\hat{A} \le 0$), we retain the standard DAPO clipping mechanism as a critical structural safeguard. We visualize the Cap for both DAPO and UP-DAPO under $\hat{A} \le 0$ in \textbf{Fig.\,\ref{fig:prob_capacity_comparison}(c)}. Detailed derivations are provided in \textbf{Appendix\,\ref{app:neg_cap}}. This ensures we do not execute overly aggressive penalization updates on wrong rollouts, thereby strictly preventing training instability.

The same asymmetric principle applies directly to GRPO \cite{shao2024deepseekmath}, another widely used token-level algorithm that employs symmetric clipping (a single $\epsilon$ for both bounds) and a sequence-level group-normalized advantage $\hat{A}_i$ shared across all tokens within the same rollout. \textbf{UP-GRPO} replaces the clipped update for positive advantages with the unbounded log-policy objective, while retaining the standard GRPO clipping safeguard and KL penalty for negative advantages:
\begin{equation}
\begin{aligned}
\mathcal{J}_{\text{UP-GRPO}}(\theta) &= \mathbb{E}_{q \sim \mathcal{Q}, \{o_i\}_{i=1}^G \sim \pi_{\text{old}}} \Bigg[ \frac{1}{G} \sum_{i=1}^G \frac{1}{|o_i|} \sum_{t=1}^{|o_i|} \\
&\quad \begin{cases}
    \hat{A}_i \log \pi_\theta(o_{i,t}|q, o_{i,<t}) - \beta \mathbb{D}_{\text{KL}}(\pi_\theta \| \pi_{\text{ref}}) & \text{if } \hat{A}_i > 0 \\
    \min \left( r_{i,t}(\theta) \hat{A}_i, \text{clip}(r_{i,t}(\theta), 1 - \epsilon, 1 + \epsilon) \hat{A}_i \right) - \beta \mathbb{D}_{\text{KL}}(\pi_\theta \| \pi_{\text{ref}}) & \text{if } \hat{A}_i \le 0
\end{cases} \Bigg],
\end{aligned}
\label{eq:up_grpo_obj}
\end{equation}
where $\hat{A}_i$ is the group-normalized advantage shared across all tokens of rollout $o_i$, and the negative branch retains the original symmetric GRPO clipping and KL penalty to prevent training instability.

Furthermore, to demonstrate its universality across different optimization granularities, our UP framework readily extends to sequence-level algorithms, yielding \textbf{UP-GSPO}. Using the length-normalized sequence-level importance ratio $s_i(\theta) = \left( \frac{\pi_\theta(o_i|q)}{\pi_{\text{old}}(o_i|q)} \right)^{\frac{1}{|o_i|}}$, the overall UP-GSPO objective is formulated as:
\begin{equation} \label{eq:up_gspo_obj}
\mathcal{J}_{\text{UP-GSPO}}(\theta) = \mathbb{E}_{q \sim \mathcal{Q}, \{o_i\}_{i=1}^G \sim \pi_{\text{old}}} \left[ \frac{1}{G} \sum_{i=1}^G \begin{cases} 
    \hat{A}_i \left( \frac{1}{|o_i|} \sum_{t=1}^{|o_i|} \log \pi_\theta(o_{i,t}|q, o_{i,<t}) \right) & \text{if } \hat{A}_i > 0 \\
    \min \Big( s_i(\theta) \hat{A}_i, \, \text{clip}(s_i(\theta), 1 - \epsilon, 1 + \epsilon) \hat{A}_i \Big) & \text{if } \hat{A}_i \le 0
\end{cases} \right].
\end{equation}
The exact analytical derivation, which rigorously proves that the positive branch of UP-GSPO mathematically equates to an unclipped, length-normalized REINFORCE gradient, is detailed in \textbf{Appendix\,\ref{app:up_gspo_derivation}}.

\section{Experiments}
\label{sec: exp}

\subsection{Experimental Setup}
\label{subsec:experimental_setup}

\noindent \textbf{Models.} 
We evaluate the UP framework across diverse model families \cite{yang2025qwen3}, including dense LLMs in both base (Qwen3-14B-Base) and instruct (Qwen3-8B) versions, an MoE model (Qwen3-30B-A3B-Base), and a vision-language model (Qwen3-VL-8B-Instruct).

\noindent \textbf{Training and Evaluation.} 
We adopt three protocols: (i) following \citet{yu2025dapo}, we train on DAPO-17K-MATH and evaluate on AIME24 \cite{codeforcesamerican}, reporting average accuracy (Avg@32) and majority voting accuracy (Maj@32) over 32 sampled trajectories, together with Best@32 (the probability of generating at least one correct answer within 32 samples); (ii) following \citet{liu2025understanding}, we train on MATH (Levels 3-5) \cite{lightman2023let} and evaluate Pass@1 on AIME24 \cite{codeforcesamerican}, AMC23, MATH500, Minerva \cite{lewkowycz2022solving}, and OlympiadBench \cite{he2024olympiadbench}; (iii) following \citet{zhao2025geometric}, we train on the Geometry3K \cite{lu2021inter} training set and evaluate Pass@1 on the Geometry3K test set.

\noindent \textbf{Algorithms and Baselines.} 
We instantiate three UP variants under our framework: \textbf{UP-GRPO}, \textbf{UP-DAPO}, and \textbf{UP-GSPO}. We compare against twelve representative RL baselines: DAPO \cite{yu2025dapo}, GRPO \cite{shao2024deepseekmath}, GMPO \cite{zhao2025geometric}, ASPO \cite{wang2025aspo}, CISPO \cite{chen2025minimax}, Dr.\,GRPO \cite{liu2025understanding}, W-REINFORCE \cite{zhu2025surprising}, REINFORCE++ \cite{hu2025reinforce++}, RLOO \cite{ahmadian2024back}, DPPO \cite{qi2026rethinking}, GSPO \cite{zheng2025group}, and SAPO \cite{gao2025soft}.

\noindent \textbf{Implementation Details.} 
Our implementation is based on the verl framework \cite{sheng2025hybridflow}, utilizing vLLM for efficient rollout generation and evaluation \cite{kwon2023efficient}. Comprehensive details are provided in \textbf{Appendix\,\ref{app:details}}.

\subsection{Performance, Exploration, and Stability Analysis of UP-DAPO}
\label{subsec:up_dapo_analysis}

\begin{figure}[htbp]
    \centering
    \begin{subfigure}{0.47\textwidth}
        \centering
        \includegraphics[width=\textwidth]{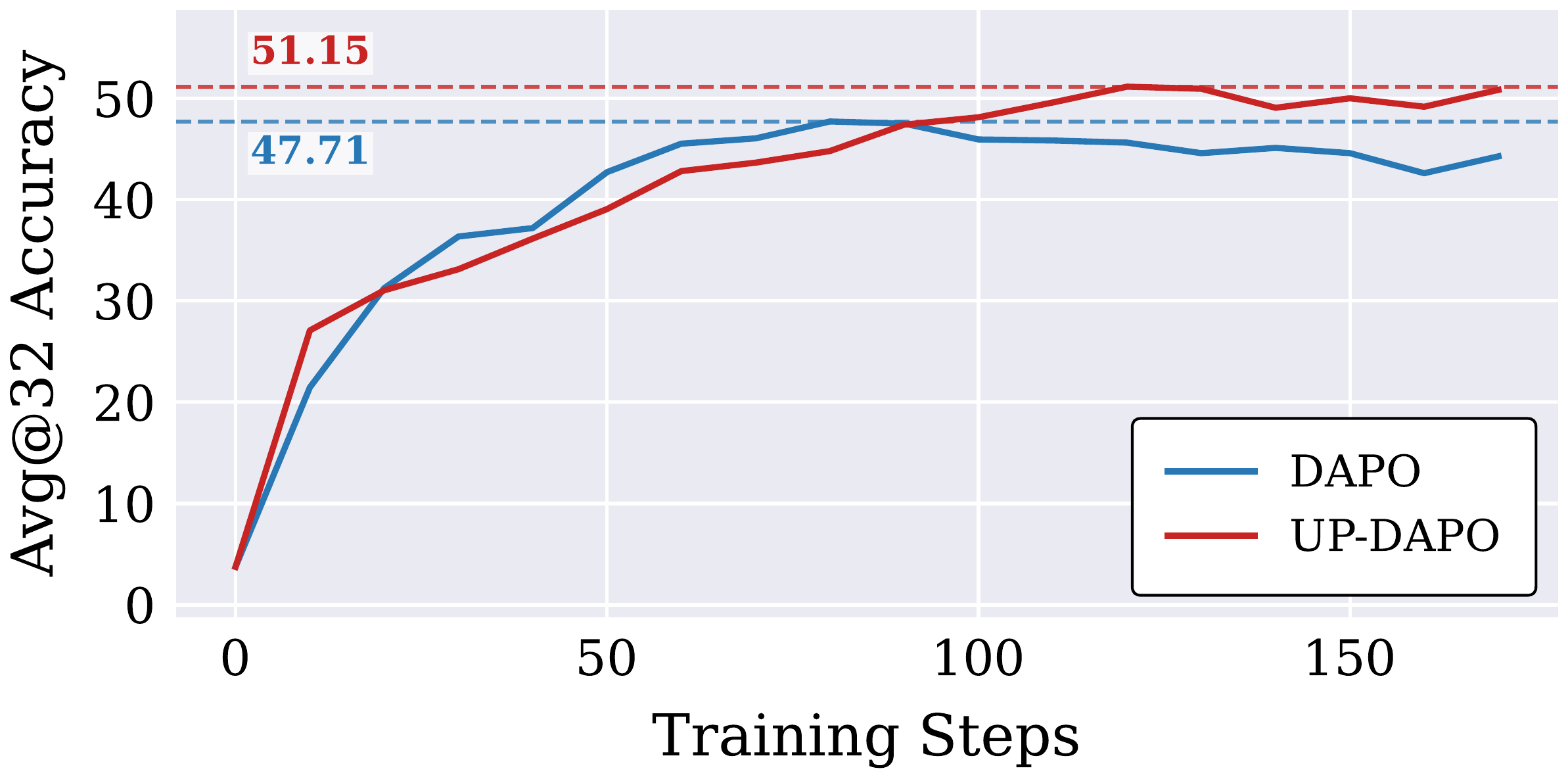}
        \caption{Avg@32}
        \label{fig:avg32_plot}
    \end{subfigure}
    \hfill
    \begin{subfigure}{0.47\textwidth}
        \centering
        \includegraphics[width=\textwidth]{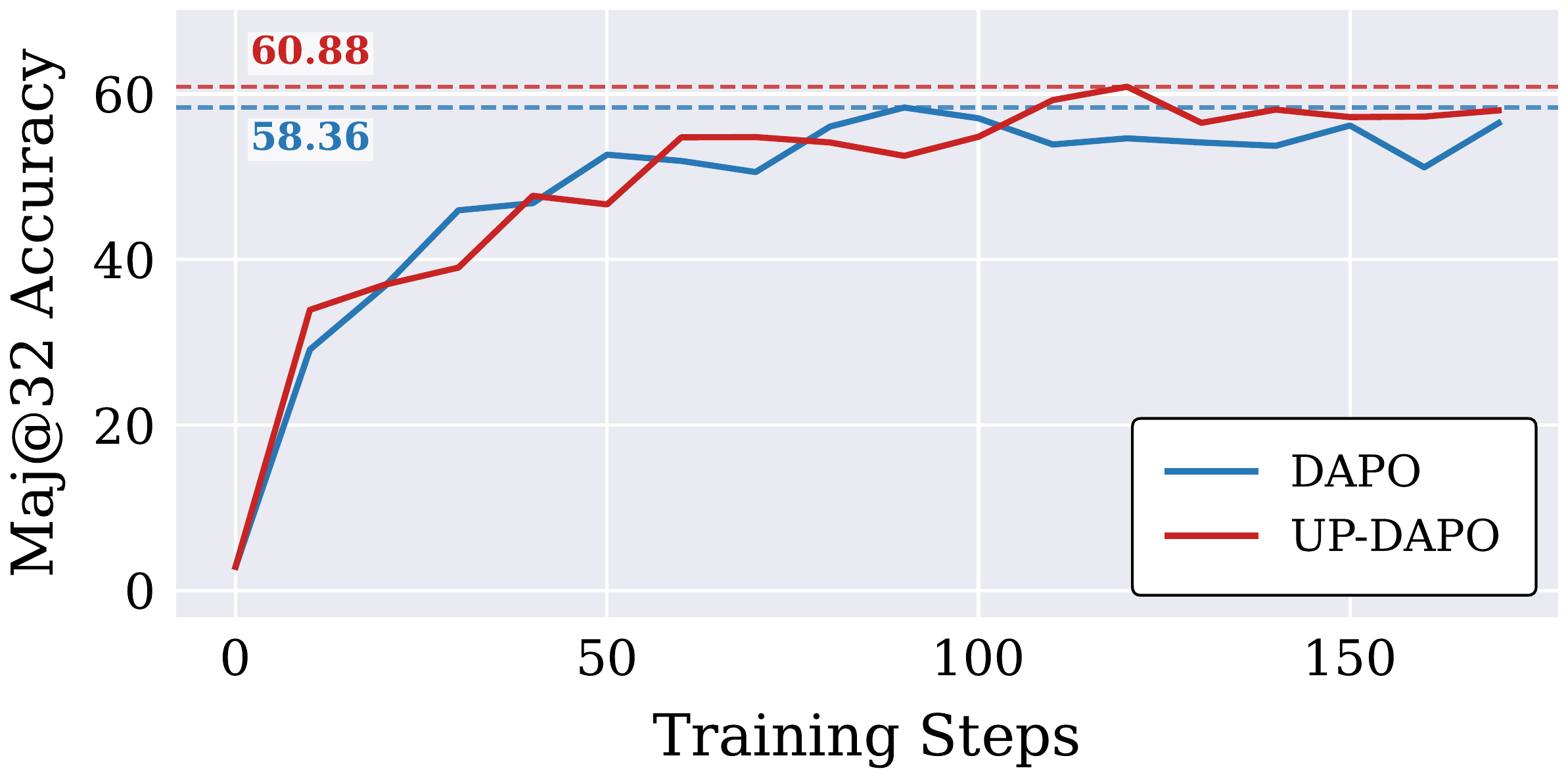}
        \caption{Maj@32}
        \label{fig:maj32_plot}
    \end{subfigure}
    \caption{Performance comparison of DAPO and UP-DAPO on Qwen3-14B-Base during training: (a) Avg@32 and (b) Maj@32 accuracy on AIME24. Peak performance for each method is denoted by dashed lines and colored labels.}
    \label{fig:aime24_performance}
\end{figure}

\noindent \textbf{Evaluation on AIME24 during Training.} 
As illustrated in \textbf{Fig.\,\ref{fig:aime24_performance}}, we monitor the performance evolution of DAPO and \textbf{UP-DAPO} on the AIME24 evaluation set throughout the training trajectory. The results in Fig.\,\ref{fig:aime24_performance}(a) and Fig.\,\ref{fig:aime24_performance}(b) show that UP-DAPO achieves superior performance over the standard DAPO baseline as training progresses. Specifically, UP-DAPO achieves a peak Avg@32 of 51.15, significantly higher than DAPO's 47.71. For Maj@32, UP-DAPO reaches 60.88, surpassing DAPO's 58.36. Notably, the evaluation curves show that UP-DAPO maintains an upward momentum and establishes a clear performance gap in the later stages of optimization, suggesting that the removal of the $\pi_{\text{old}}$ anchor allows for more effective policy optimization in reasoning tasks.

\begin{figure}[htbp]
    \centering
    \begin{subfigure}{0.47\textwidth}
        \centering
        \includegraphics[width=\textwidth]{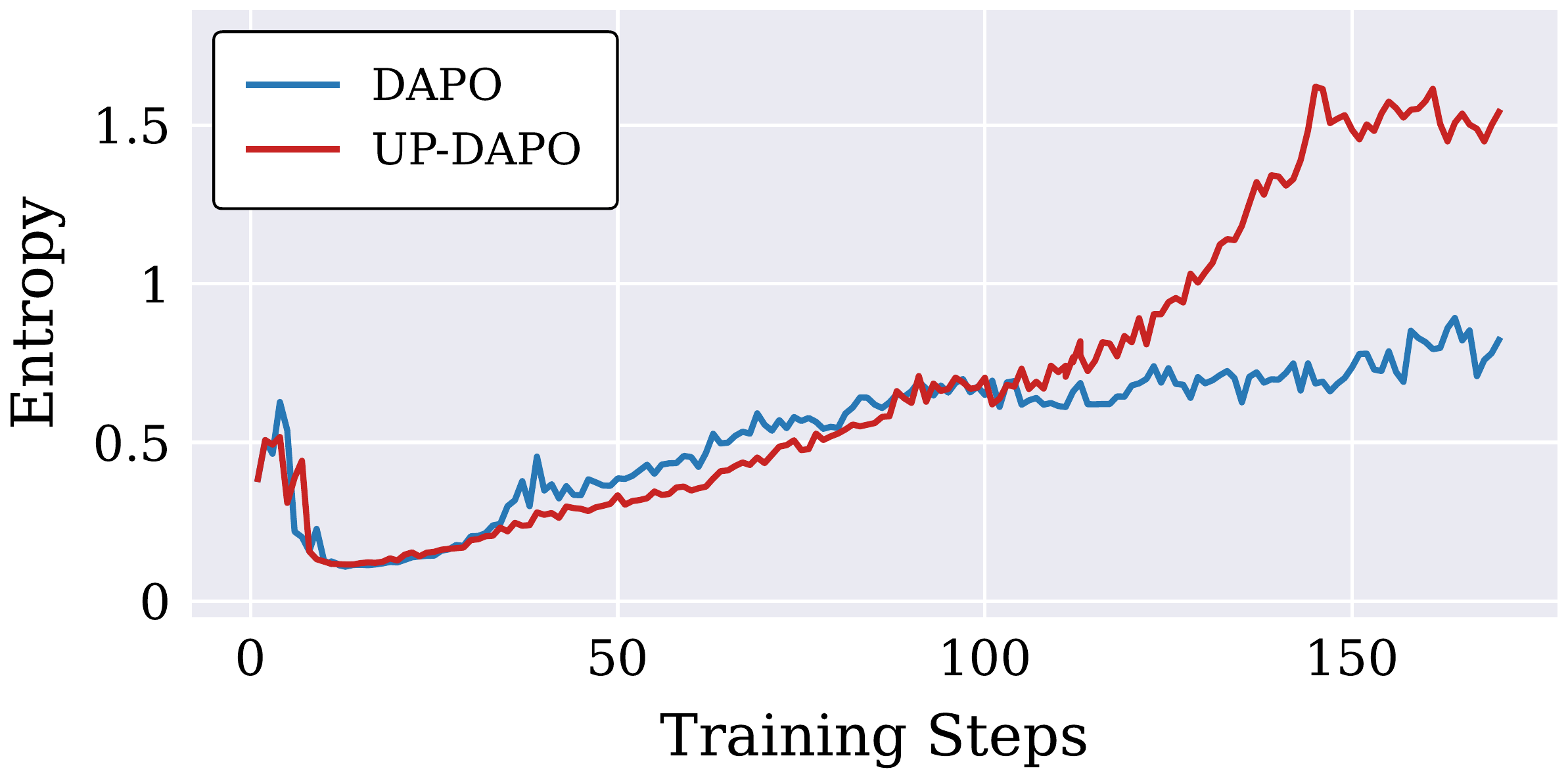}
        \caption{Entropy}
        \label{fig:entropy_plot}
    \end{subfigure}
    \hfill
    \begin{subfigure}{0.47\textwidth}
        \centering
        \includegraphics[width=\textwidth]{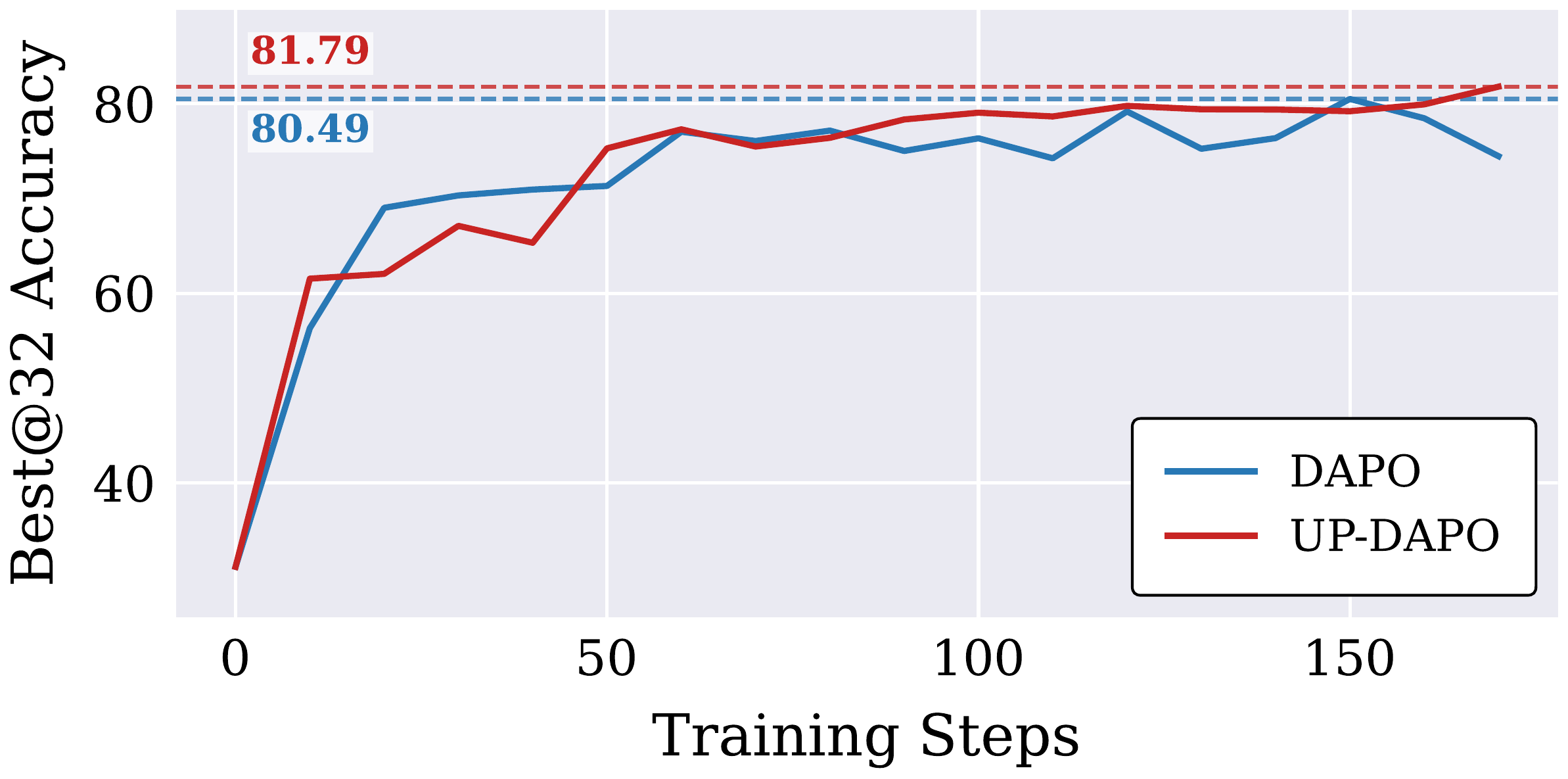}
        \caption{Best@32}
        \label{fig:best32_plot}
    \end{subfigure}
    \caption{Exploration capacity of DAPO and UP-DAPO on Qwen3-14B-Base during training: (a) entropy of generation probabilities on DAPO-17K-MATH and (b) Best@32 accuracy on AIME24. Peak performance for each method is denoted by dashed lines and colored labels.}
    \label{fig:exploration_capacity}
\end{figure}

\noindent \textbf{Enhancement of Exploration Capacity.} 
In \textbf{Fig.\,\ref{fig:exploration_capacity}(a)}, we present the entropy of generation probabilities for UP-DAPO and DAPO on the DAPO-17K-MATH training set to evaluate their respective exploration capacities. It is evident that UP-DAPO exhibits higher entropy compared to DAPO, which directly correlates to a maximized exploration capacity during the training process. To further validate this, we report the Best@32 accuracy on the AIME24 evaluation set throughout the training trajectory in \textbf{Fig.\,\ref{fig:exploration_capacity}(b)}. The results show that UP-DAPO achieves a peak Best@32 of 81.79, surpassing DAPO's 80.49. This improvement in the performance upper bound confirms that the increased entropy in UP-DAPO successfully translates into a stronger exploration capability, allowing the model to discover higher-quality reasoning paths.

\begin{figure}[htbp]
    \centering
    \begin{subfigure}{0.47\textwidth}
        \centering
        \includegraphics[width=\textwidth]{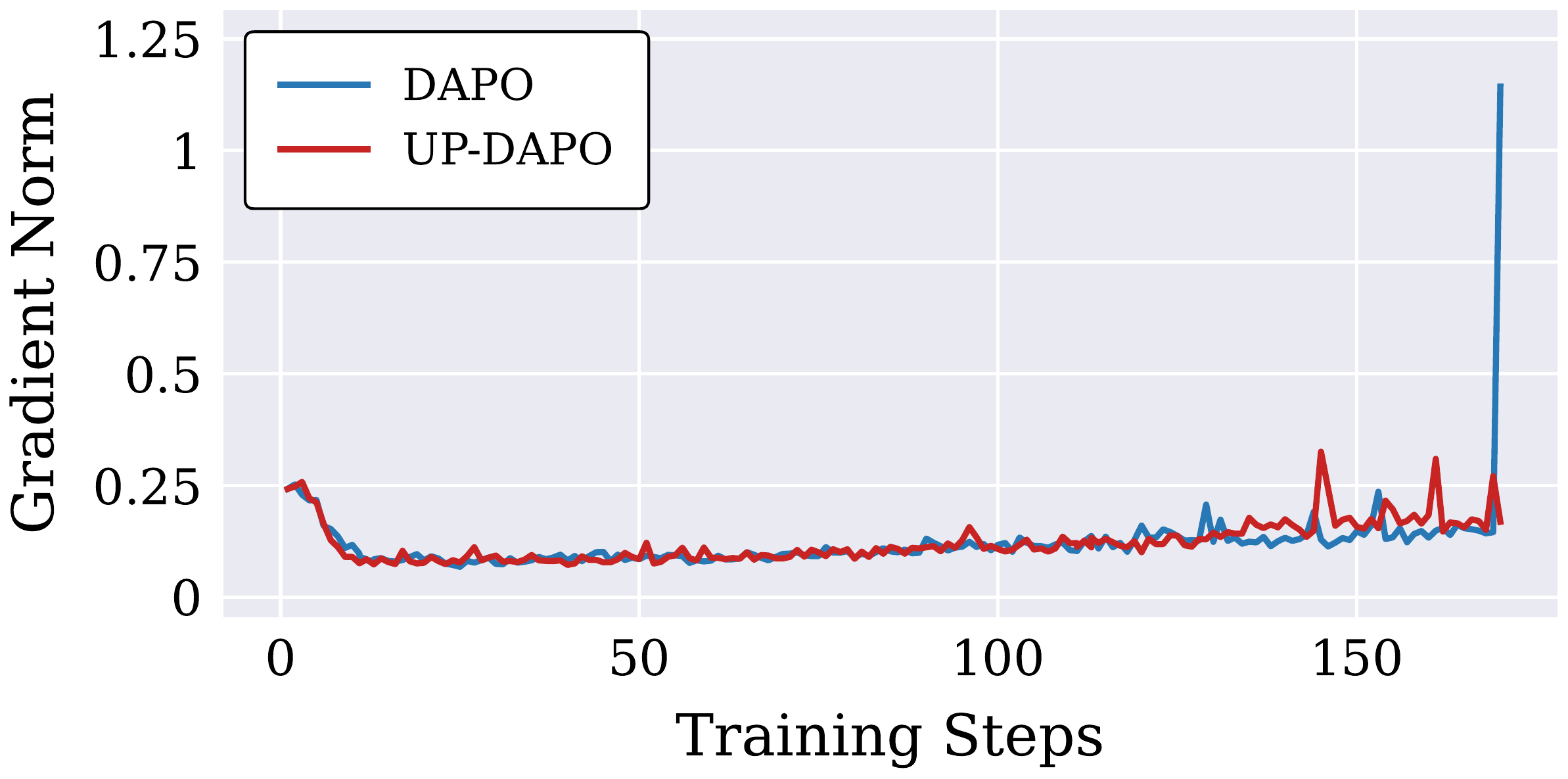}
        \caption{Gradient norm}
        \label{fig:grad_norm_plot}
    \end{subfigure}
    \hfill
    \begin{subfigure}{0.47\textwidth}
        \centering
        \includegraphics[width=\textwidth]{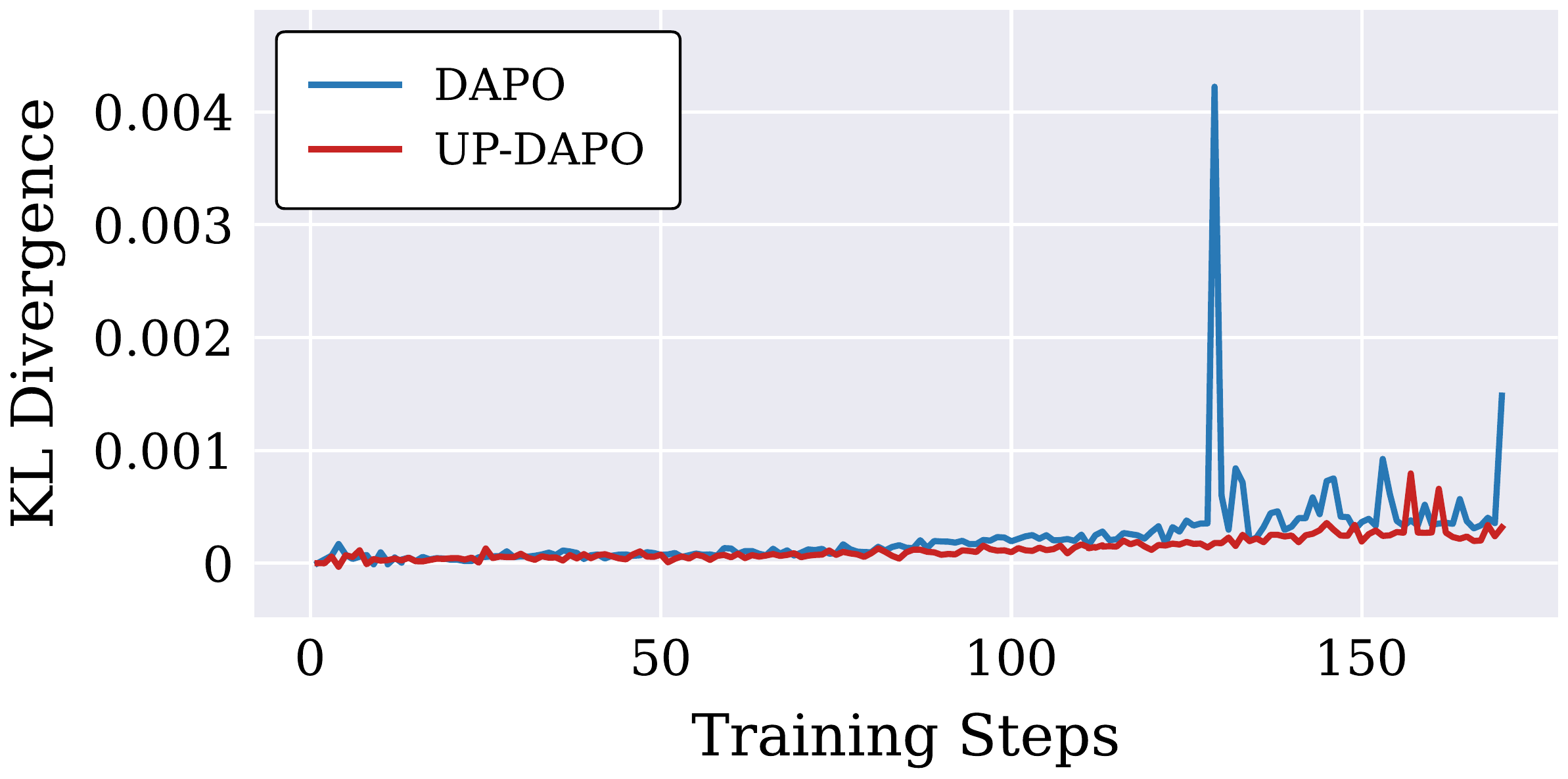}
        \caption{KL divergence}
        \label{fig:kl_plot}
    \end{subfigure}
    \caption{Training stability of DAPO and UP-DAPO on Qwen3-14B-Base during training: (a) gradient norm and (b) KL divergence between the active policy and the reference model on DAPO-17K-MATH.}
    \label{fig:training_stability}
\end{figure}

\noindent \textbf{Analysis of Training Stability.} 
In \textbf{Fig.\,\ref{fig:training_stability}}, we evaluate the stability of the training process. Despite the increased exploration observed in previous metrics, the gradient norm in \textbf{Fig.\,\ref{fig:training_stability}(a)} and the KL divergence in \textbf{Fig.\,\ref{fig:training_stability}(b)} of UP-DAPO remain comparable to, or even slightly lower than, those of the standard DAPO baseline. This observation is critical as it demonstrates that although UP-DAPO removes the importance sampling and clipping mechanisms for correct rollouts to maximize exploration, it does so without sacrificing stability or causing the policy to deviate from the reference model. By maintaining these metrics within a stable range, our method effectively prevents training instability while preserving a more expansive exploration capacity.

\subsection{Ablation Studies of UP-DAPO}
\label{subsec:ablation}

\begin{figure}[htbp]
    \centering
    \begin{subfigure}{0.47\textwidth}
        \centering
        \includegraphics[width=\textwidth]{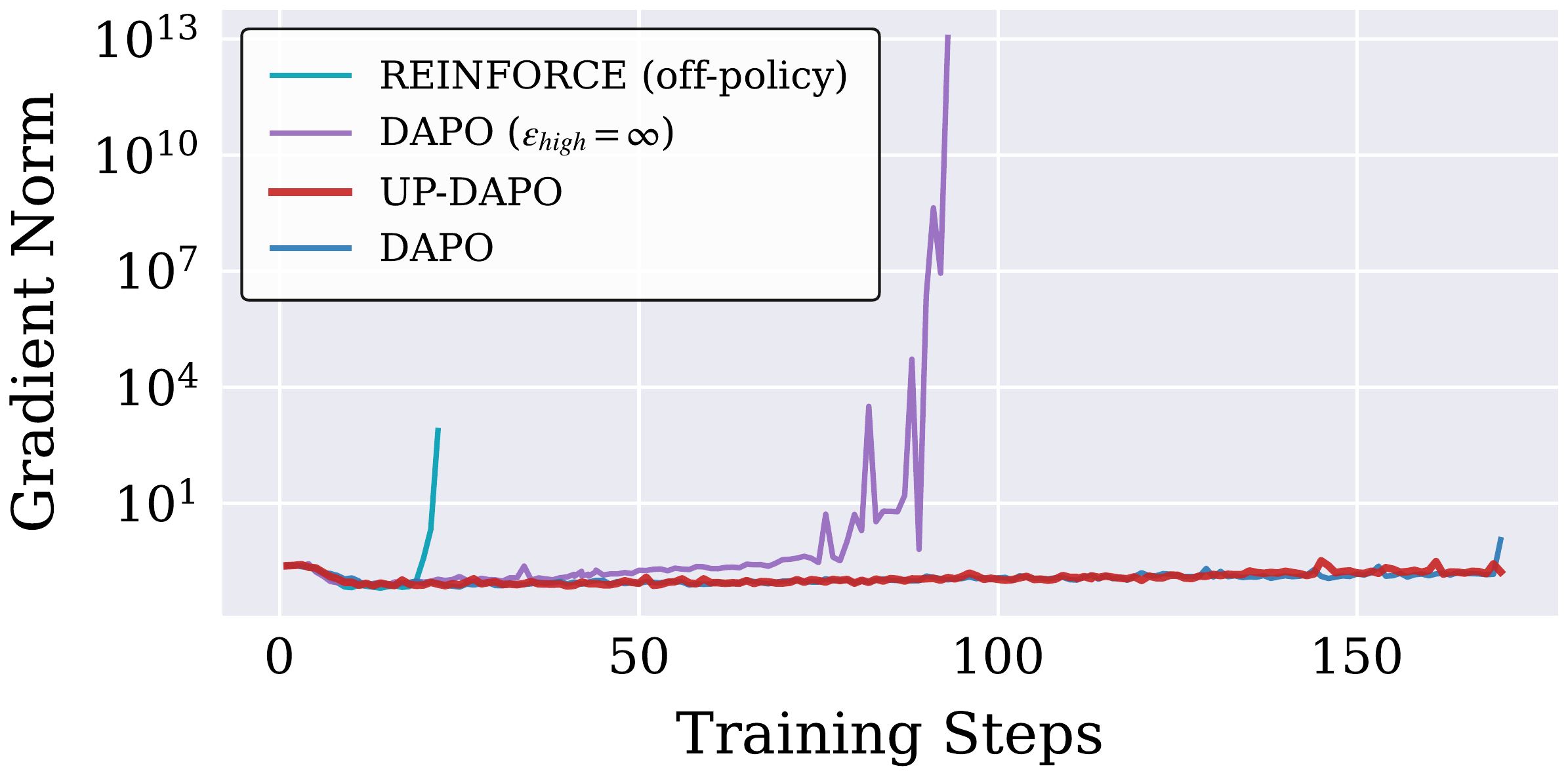}
        \caption{Gradient norm}
        \label{fig:grad_norm_log_plot}
    \end{subfigure}
    \hfill
    \begin{subfigure}{0.47\textwidth}
        \centering
        \includegraphics[width=\textwidth]{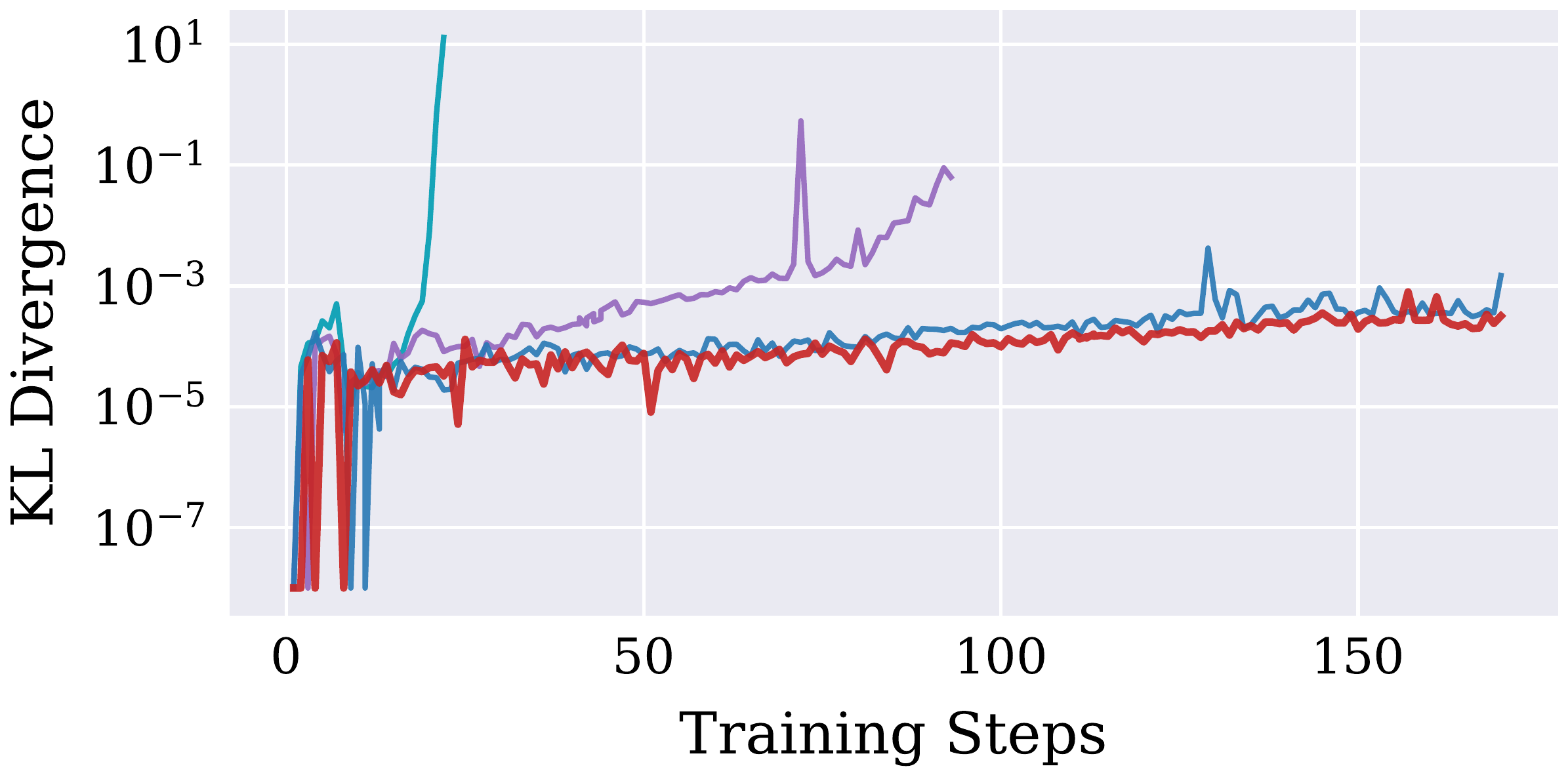}
        \caption{KL divergence}
        \label{fig:kl_log_plot}
    \end{subfigure}
    \caption{Training stability of DAPO, DAPO ($\epsilon_{\text{high}} = \infty$), REINFORCE (off-policy), and UP-DAPO on Qwen3-14B-Base during training: (a) gradient norm and (b) KL divergence between the active policy and the reference model on DAPO-17K-MATH.}
    \label{fig:ablation_study}
\end{figure}

\noindent \textbf{Necessity of Self-Anchored Ratio.} 
We first examine whether the standard DAPO framework can achieve comparable exploration by simply relaxing its constraints within the original importance sampling ratio $r_{i,t}(\theta)$. This experiment is designed to demonstrate the necessity of our self-anchored ratio $\tilde{r}_{i,t}(\theta)$ in \textbf{Eq.\,\ref{eq:up_obj}}. We implement a modified DAPO baseline that sets the upper clip bound to infinity ($\epsilon_{\text{high}} = \infty$). As shown by the purple curves in \textbf{Fig.\,\ref{fig:ablation_study}}, while this modification initially remains stable, it induces severe training instability beyond 80 steps. The gradient norm explodes to $10^{13}$, accompanied by a surge in KL divergence. This instability demonstrates that merely removing the upper clip bound while still anchoring the ratio to $\pi_{\text{old}}$ is fundamentally insufficient for safe exploration. Our findings highlight the critical role of the stop-gradient operator: by replacing $\pi_{\text{old}}$ with $\text{sg}(\pi_\theta)$ to formulate the self-anchored modified ratio $\tilde{r}_{i,t}(\theta)$, we effectively eliminate the root cause of instability and safely enable unbounded optimization.

\noindent \textbf{Necessity of Asymmetric Objective Design.} 
We further investigate the structural necessity of the asymmetric mechanism in \textbf{Eq.\,\ref{eq:up_dapo_obj}} by evaluating a symmetric, unbounded baseline. This variant adopts an off-policy REINFORCE-style formulation that applies unbounded updates to both positive and negative advantages, effectively removing all clipping constraints. As shown by the cyan curves in Fig.\,\ref{fig:ablation_study}, applying the unbounded mechanism to wrong rollouts leads to immediate and catastrophic training instability within the first 25 training steps. Both the gradient norm and KL divergence exhibit an uncontrolled vertical surge. This confirms that while the unbounded formulation is beneficial for correct rollouts, DAPO's clipping remains a strictly indispensable safeguard for wrong rollouts to prevent training instability.

\subsection{Comparison with Other RL Baselines}
\label{subsec:baseline_comparison}

\begin{table}[htbp]
\centering
\caption{Performance of UP-GRPO against eleven RL baselines on AIME24, AMC23, MATH500, Minerva and OlympiadBench. All algorithms are trained on MATH (Levels 3-5) using Qwen3-8B following the protocol of \citet{liu2025understanding}. Performance is evaluated by Pass@1 accuracy ($\%$) on each benchmark and the \textbf{Average} across the five benchmarks. The best score in each column is in \textbf{bold}.}
\label{tab:baseline_comparison}
\setlength{\tabcolsep}{5pt}
\begin{tabular}{lccccc|c}
\toprule
\textbf{Method} & \textbf{AIME24} & \textbf{AMC23} & \textbf{MATH500} & \textbf{Minerva} & \textbf{OlympiadBench} & \textbf{Average} \\
\midrule
GRPO \cite{shao2024deepseekmath}            & 35.73          & 75.00          & 86.00          & 30.88          & 51.34          & 55.79 \\
Dr.\,GRPO \cite{liu2025understanding}       & 33.33          & 85.00          & 85.80          & 30.15          & 51.19          & 57.09 \\
CISPO \cite{chen2025minimax}                & 38.02          & \textbf{87.50} & 86.60          & 29.04          & 55.65          & 59.36 \\
DPPO \cite{qi2026rethinking}                & 40.10          & \textbf{87.50} & 86.20          & 30.51          & 53.27          & 59.52 \\
GMPO \cite{zhao2025geometric}               & 37.50          & \textbf{87.50} & 87.00          & \textbf{31.25} & 55.06          & 59.66 \\
GSPO \cite{zheng2025group}                  & 40.52          & 85.00          & 88.20          & \textbf{31.25} & 55.80          & 60.15 \\
SAPO \cite{gao2025soft}                     & 39.90          & 82.50          & 87.20          & 30.88          & 55.65          & 59.23 \\
REINFORCE++ \cite{hu2025reinforce++}        & 20.52          & 62.50          & 78.80          & 29.78          & 42.26          & 46.77 \\
RLOO \cite{ahmadian2024back}                & 31.67          & 80.00          & 85.20          & 28.68          & 50.60          & 55.23 \\
W-REINFORCE \cite{zhu2025surprising}        & 35.52          & 80.00          & 85.80          & 30.15          & 53.27          & 56.95 \\
ASPO \cite{wang2025aspo}                    & 37.50          & 85.00          & 87.60          & 29.78          & \textbf{58.48} & 59.67 \\
\midrule
\rowcolor{blue!10}
\textbf{UP-GRPO (Ours)}                     & \textbf{41.04} & \textbf{87.50} & \textbf{88.40} & \textbf{31.25} & 58.33          & \textbf{61.31} \\
\bottomrule
\end{tabular}
\end{table}

\noindent \textbf{Overall Performance.} 
We benchmark \textbf{UP-GRPO} against eleven representative RL baselines under the unified protocol of \citet{liu2025understanding}, training on MATH (Levels 3-5) with Qwen3-8B and evaluating Pass@1 accuracy across five reasoning benchmarks. As reported in \textbf{Table\,\ref{tab:baseline_comparison}}, UP-GRPO attains the best average accuracy of 61.31\%, surpassing all competing baselines including the strongest prior method GSPO (60.15\%) by an absolute margin of 1.16\%. Beyond the average, UP-GRPO ranks first or ties for first on four out of the five individual benchmarks, achieving 41.04\% on AIME24, 87.50\% on AMC23, 88.40\% on MATH500, and 31.25\% on Minerva, while remaining highly competitive on OlympiadBench (58.33\%, second only to ASPO's 58.48\%). These consistent gains across benchmarks of varying difficulty demonstrate that the unbounded positive update yields a broadly effective, rather than benchmark-specific, improvement over the GxPO family.

\begin{figure}[htbp]
    \centering
    % 第一行：图例（不加caption，因此不会占用编号 (a)）
    \begin{subfigure}{0.95\textwidth}
        \centering
        \includegraphics[width=\textwidth]{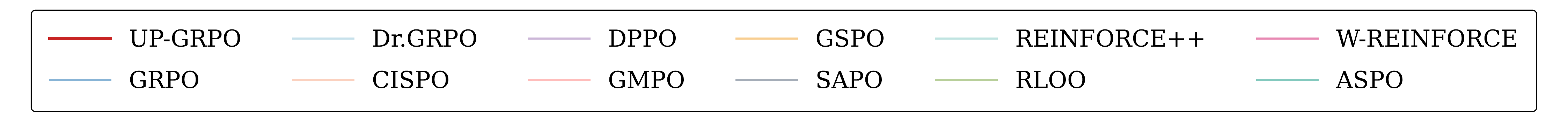}
    \end{subfigure}
    
    % \vspace{-4mm} % 调整图例和下方子图的间距
    % 此处的一个空行非常重要，它起到了换行（分段）的作用
    
    % 第二行左侧：Entropy 图 -> 自动编号为 (a)
    \begin{subfigure}{0.47\textwidth}
        \centering
        \includegraphics[width=\textwidth]{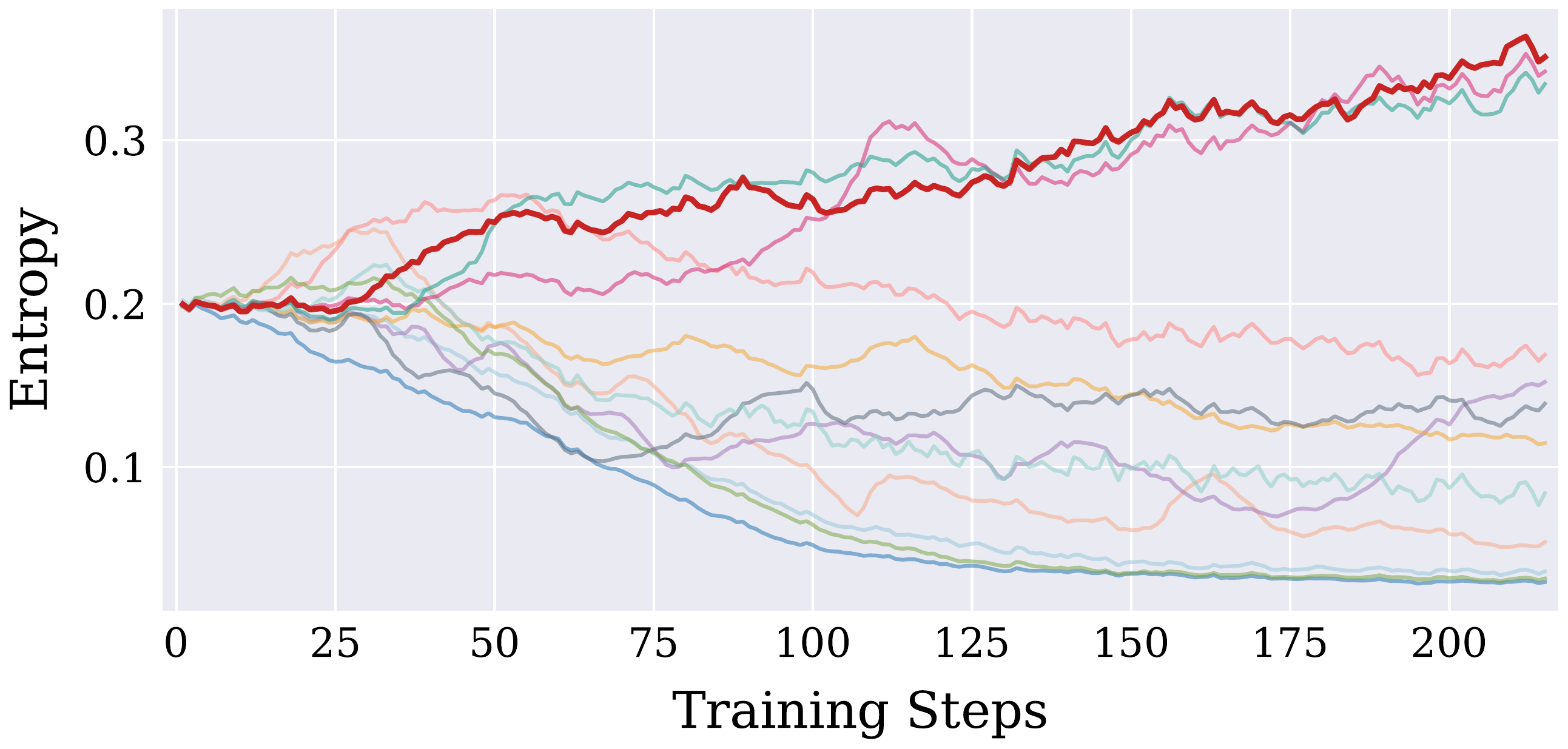}
        \caption{Entropy}
        \label{fig:grpo_entropy_plot}
    \end{subfigure}
    \hfill
    % 第二行右侧：KL divergence 图 -> 自动编号为 (b)
    \begin{subfigure}{0.47\textwidth}
        \centering
        \includegraphics[width=\textwidth]{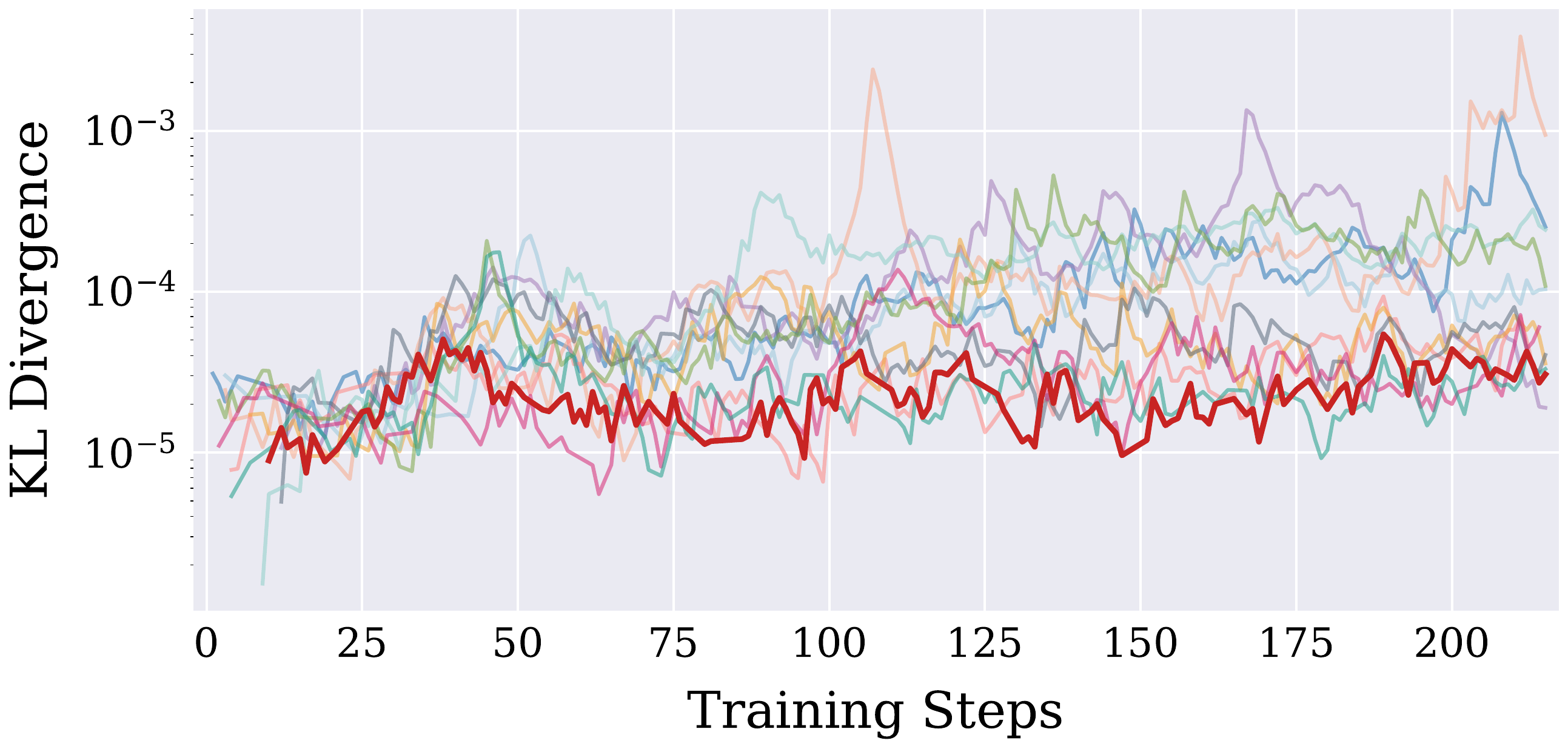}
        \caption{KL divergence}
        \label{fig:grpo_kl_log_plot}
    \end{subfigure}
    
    \caption{Exploration capacity and training stability of UP-GRPO against eleven RL baselines on Qwen3-8B during training: (a) Entropy and (b) KL divergence between the active policy and the reference model on MATH (Levels 3-5).}
    \label{fig:grpo_family_comparison}
\end{figure}

\noindent \textbf{Exploration Capacity and Training Stability.} 
To understand the source of these gains, we further examine the exploration capacity (measured by policy entropy) and the training stability (measured by KL divergence between the active policy and the reference model) of all twelve methods throughout training. \textbf{Fig.\,\ref{fig:grpo_family_comparison}(a)} reveals a clear bifurcation among the baselines. The majority of methods, including GRPO, Dr.\,GRPO, CISPO, DPPO, GMPO, GSPO, SAPO, REINFORCE++, and RLOO, exhibit a steady entropy collapse as training proceeds, indicating that their policies progressively lose the ability to explore alternative reasoning trajectories. In contrast, UP-GRPO maintains a consistently rising entropy throughout training, confirming that the unbounded positive update preserves and even amplifies exploration capacity. Notably, the only other methods that avoid entropy collapse are W-REINFORCE and ASPO, both of which—like UP-GRPO—adopt asymmetric treatments of correct and wrong rollouts. This shared behavior reaffirms that asymmetric optimization is a necessary structural ingredient for sustaining exploration in long-horizon RL training. \textbf{Fig.\,\ref{fig:grpo_family_comparison}(b)} further shows that UP-GRPO's KL divergence remains among the lowest across the entire training trajectory, indicating that the active policy stays close to the reference model and thereby ensures stable optimization. Taken together, these results show that UP-GRPO simultaneously achieves the strongest exploration capacity and one of the most stable training dynamics, providing a principled explanation for its superior downstream performance.

\subsection{Universality of UP across Algorithms, Architectures, and Modalities}
\label{subsec:universality}

\begin{figure}[htbp]
    \centering
    \begin{subfigure}{0.47\textwidth}
        \centering
        \includegraphics[width=\textwidth]{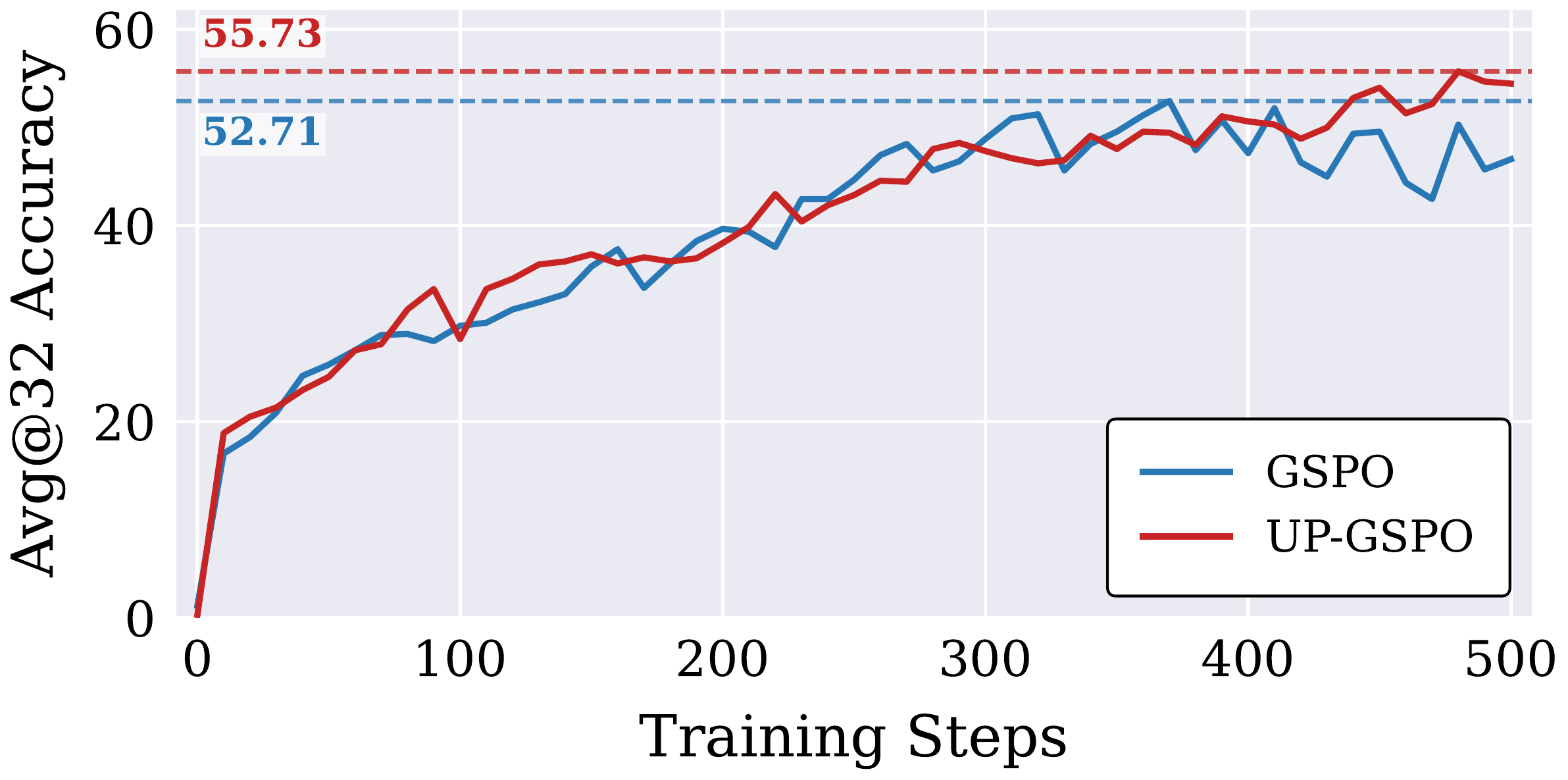}
        \caption{Avg@32}
    \end{subfigure}
    \hfill
    \begin{subfigure}{0.47\textwidth}
        \centering
        \includegraphics[width=\textwidth]{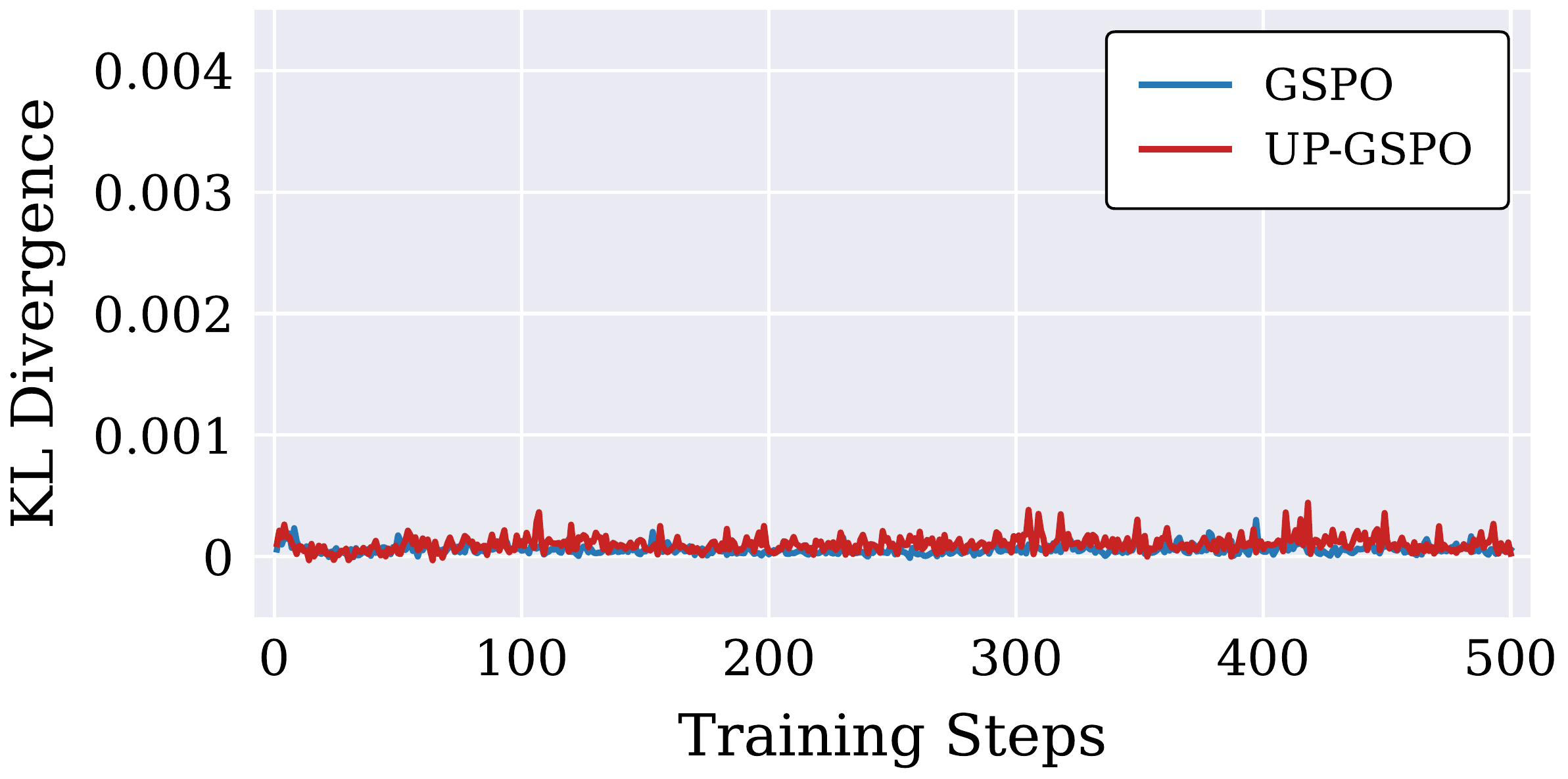}
        \caption{KL divergence}
    \end{subfigure}
    \caption{Performance and stability comparison of GSPO and UP-GSPO on Qwen3-30B-A3B-Base during training: (a) Avg@32 accuracy on AIME24 and (b) KL divergence between the active policy and the reference model on DAPO-17K-MATH. Peak performance for each method is denoted by dashed lines and colored labels.}
    \label{fig:gspo}
\end{figure}

\noindent \textbf{Universality across GxPO Variants and Model Architectures.} 
To validate that the UP framework serves as a universal plugin across diverse GxPO algorithms and model architectures, we evaluate \textbf{UP-GSPO} on the Qwen3-30B-A3B-Base MoE model. This experiment tests the framework's adaptability to the architectural shift from Dense to MoE. As illustrated in \textbf{Fig.\,\ref{fig:gspo}}, UP-GSPO consistently achieves superior performance compared to the standard GSPO baseline. Specifically, UP-GSPO reaches a peak Avg@32 accuracy of 55.73\%, providing an absolute improvement of 3.02\% over GSPO's 52.71\%. Crucially, the KL divergence of UP-GSPO remains nearly identical to the baseline throughout the training process. This demonstrates that our design in \textbf{Eq.\,\ref{eq:up_gspo_obj}} effectively resolves the exploration-stability dilemma, regardless of the optimization granularity (from token-level DAPO to sequence-level GSPO) or the model architecture (from a dense model to MoE). These results confirm that UP is a robust enhancement, capable of safely increasing exploration for correct reasoning trajectories while preventing training instability across the broader GxPO family and diverse architectures.

\begin{figure}[htbp]
    \centering
    \begin{subfigure}{0.47\textwidth}
        \centering
        \includegraphics[width=\textwidth]{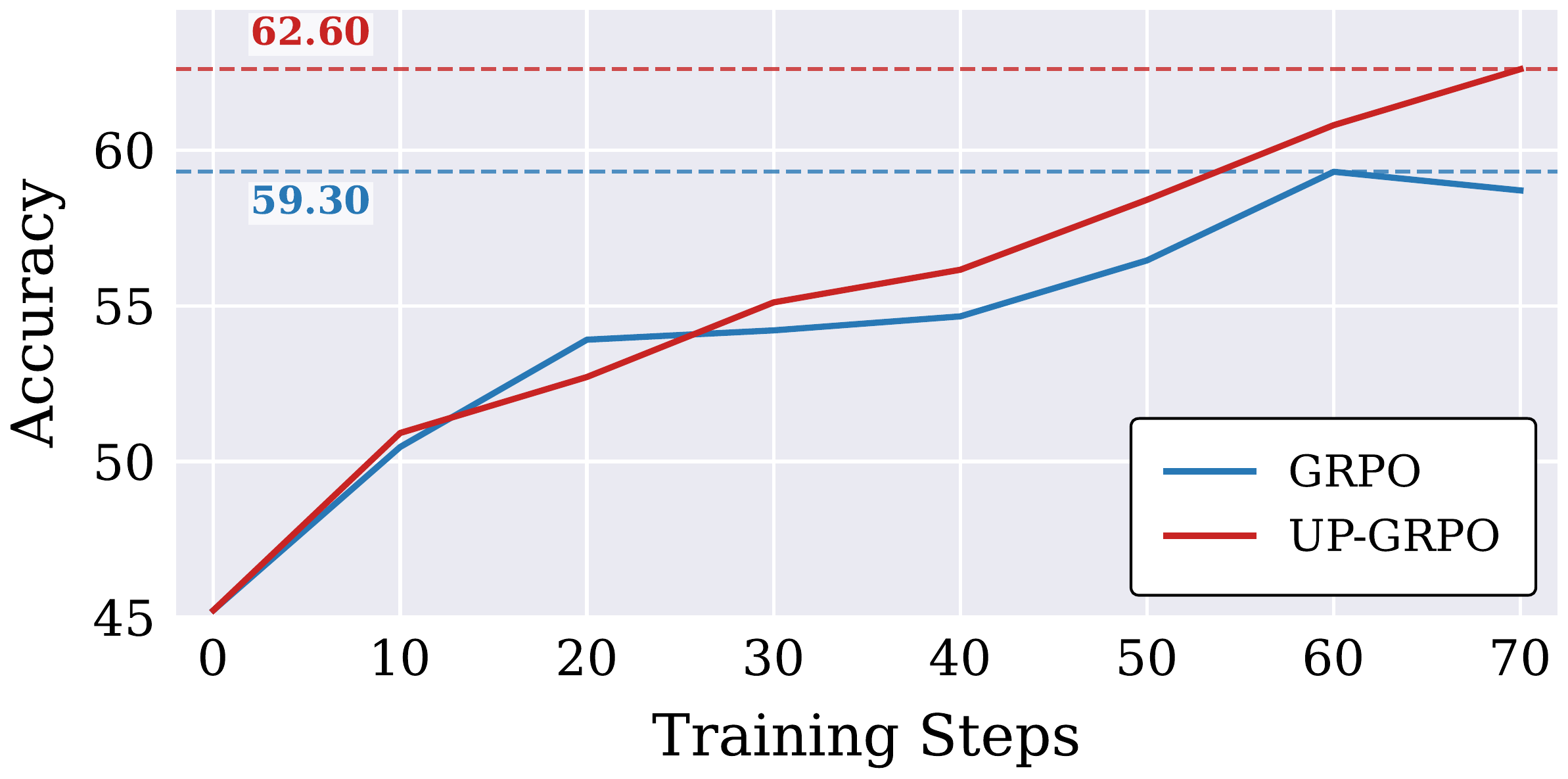}
        \caption{Accuracy}
        \label{fig:vlm_avg32_plot}
    \end{subfigure}
    \hfill
    \begin{subfigure}{0.47\textwidth}
        \centering
        \includegraphics[width=\textwidth]{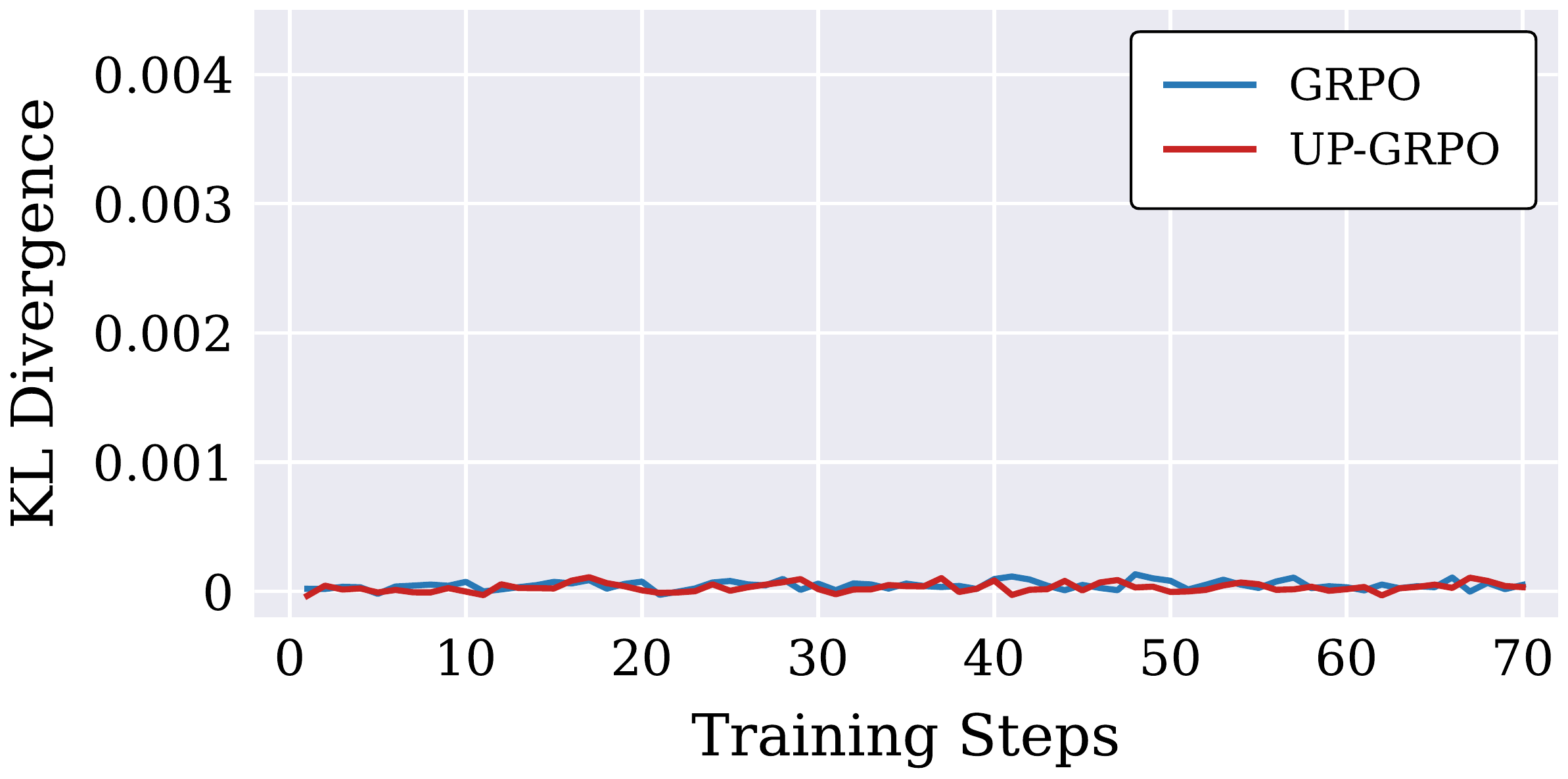}
        \caption{KL divergence}
        \label{fig:vlm_kl_plot}
    \end{subfigure}
    \caption{Performance and stability comparison of GRPO and UP-GRPO on Qwen3-VL-8B-Instruct during training: (a) accuracy on Geometry3K test set and (b) KL divergence between the active policy and the reference model. Peak performance for each method is denoted by dashed lines and colored labels.}
    \label{fig:vlm_performance}
\end{figure}

\noindent \textbf{Generalization to Multimodal Setting and New Training Data.}
To further validate that UP is both modality-agnostic and training-data-agnostic, we evaluate \textbf{UP-GRPO} (formulated in \textbf{Eq.\,\ref{eq:up_grpo_obj}}) against the standard GRPO baseline using the Qwen3-VL-8B-Instruct vision-language model trained on the Geometry3K \cite{lu2021inter} visual reasoning dataset. This experiment extends our validation beyond language reasoning to multimodal geometric problem solving, where the model must jointly process visual diagrams and textual descriptions.
As illustrated in \textbf{Fig.\,\ref{fig:vlm_performance}(a)}, UP-GRPO consistently achieves superior accuracy over GRPO throughout the training process. Specifically, UP-GRPO reaches a peak accuracy of 62.60\%, representing an absolute improvement of 3.30\% over GRPO's 59.30\%. As shown in \textbf{Fig.\,\ref{fig:vlm_performance}(b)}, the KL divergence of UP-GRPO remains nearly identical to that of GRPO, confirming that the unbounded positive update does not induce any additional instability even in the multimodal training regime. These results demonstrate that UP generalizes seamlessly across modalities and datasets, establishing it as a truly universal plug-and-play objective for RL-based training of both language and vision-language models.
\section{Conclusion}
\label{sec:conclusion}

To overcome the inherent exploration-stability dilemma in IS-based reinforcement learning, we have explored the role of Unbounded Positive Asymmetric Optimization (UP) in maximizing exploration capacity and established its universal applicability across both token-level and sequence-level frameworks. Through the formalization of Probability Capacity (Cap), we have demonstrated how anchoring the policy via the stop-gradient operator mathematically bypasses conservative clipping constraints for positive advantages. Extensive experiments confirm that UP significantly improves reasoning performance across diverse RL algorithms (DAPO, GSPO, and GRPO), model architectures (Dense, MoE, and vision-language), and training modalities (language and multimodal), with our asymmetric design serving as a highly effective and universal paradigm for unleashing exploration while strictly preventing training instability.

\clearpage

\bibliographystyle{plainnat}
\bibliography{main,referneces/rl}

\clearpage
\setcounter{section}{0}
\setcounter{figure}{0}
\setcounter{table}{0}
\makeatletter 
\renewcommand{\thesection}{\Alph{section}}
\renewcommand{\theHsection}{\Alph{section}}
\renewcommand{\thefigure}{A\arabic{figure}}
\renewcommand{\theHfigure}{A\arabic{figure}}
\renewcommand{\thetable}{A\arabic{table}}
\renewcommand{\theHtable}{A\arabic{table}}
\makeatother

\renewcommand{\thetable}{A\arabic{table}}
\setcounter{equation}{0}
\renewcommand{\theequation}{A\arabic{equation}}
\beginappendix

\section{Training and Evaluation Details}
\label{app:details}

In this section, we provide the comprehensive hyperparameter configurations used for training and evaluating. To ensure a strictly fair and rigorous empirical comparison, the baseline algorithms and their corresponding UP variants share identical foundational training setups, hardware allocations, and decoding strategies within their respective model classes. The only variable altered between a baseline and its UP counterpart is the specific mathematical optimization objective introduced in the main text.

\textbf{Table\,\ref{tab:training_config}} outlines the core algorithmic and optimization hyperparameters. Crucially, for both UP variants, these clipping bounds are applied asymmetrically: they function strictly as a structural safeguard for negative advantages, leaving positive advantages completely unconstrained. Consequently, the UP variants elegantly eliminate the need for the upper clip hyperparameter $\epsilon_{\text{high}}$. \textbf{Table\,\ref{tab:decoding_config}} summarizes the decoding configurations for the rollout generation phase.

\begin{table}[htbp]
\centering
\caption{Training and algorithmic hyperparameters for token-level (DAPO and UP-DAPO on Qwen3-14B-Base) and sequence-level (GSPO and UP-GSPO on Qwen3-30B-A3B-Base) optimization.}
\label{tab:training_config}
\begin{tabular}{lcccc}
\toprule
& \multicolumn{2}{c}{\textbf{Dense 14B}} & \multicolumn{2}{c}{\textbf{MoE 30B}} \\
\cmidrule(lr){2-3} \cmidrule(lr){4-5}
\textbf{Item} & \textbf{DAPO} & \textbf{UP-DAPO} & \textbf{GSPO} & \textbf{UP-GSPO} \\
\midrule
Prompt / Response max & 2,048 / 20,480 & 2,048 / 20,480 & 2,048 / 20,480 & 2,048 / 20,480 \\
Global batch size (prompts) & 512 & 512 & 256 & 256 \\
Rollout $n$ & 16 & 16 & 16 & 16 \\
PPO mini-batch size & 32 & 32 & 32 & 32 \\
% Total epochs & 1 & 1 & 10 & 10 \\
Learning rate & $1 \times 10^{-6}$ & $1 \times 10^{-6}$ & $1 \times 10^{-6}$ & $1 \times 10^{-6}$ \\
% Advantage Estimator & GRPO & GRPO & GRPO & GRPO \\
% Loss aggregation & token-mean & token-mean & sequence-level & sequence-level \\
Lower Clip $\epsilon_{\text{low}}$ & 0.2 & 0.2 & $3 \times 10^{-4}$ & $3 \times 10^{-4}$ \\
Upper Clip $\epsilon_{\text{high}}$ & 0.28 & $-$ & $4 \times 10^{-4}$ & $-$ \\
% Entropy coefficient & 0 & 0 & 0 & 0 \\
% KL penalty & disabled & disabled & disabled & disabled \\
% Parallelism & Ulysses SP=8 & Ulysses SP=8 & Megatron TP=2 & Megatron TP=2 \\
\bottomrule
\end{tabular}
\end{table}

\begin{table}[htbp]
\centering
\caption{Decoding and rollout hyperparameters for token-level (DAPO and UP-DAPO on Qwen3-14B-Base) and sequence-level (GSPO and UP-GSPO on Qwen3-30B-A3B-Base) training.}
\label{tab:decoding_config}
\begin{tabular}{lcc}
\toprule
\textbf{Item} & \textbf{DAPO / UP-DAPO} & \textbf{GSPO / UP-GSPO} \\
\midrule
Training rollout temperature & 1.0 & 1.0 \\
Training rollout Top-$p$ & 1.0 & 1.0 \\
Validation Top-$p$ & 0.7 & 0.7 \\
Top-$k$ & -1 & -1 \\
Max generation tokens & 20,480 & 20,480 \\
% vLLM GPU utilization & 0.7 & 0.4 \\
% Inference Tensor Parallel & 1 & 4 \\
\bottomrule
\end{tabular}
\end{table}

\textbf{Table\,\ref{tab:baseline_training_config}} and \textbf{Table\,\ref{tab:baseline_decoding_config}} provide the training and decoding configurations used by GRPO and UP-GRPO in the comprehensive baseline comparison on Qwen3-8B trained on MATH (level 3-5). The two methods share identical configurations; the only difference is that UP-GRPO removes the upper clipping bound for positive advantages, eliminating the need for the upper clip hyperparameter $\epsilon_{\text{high}}$. For the remaining baselines, we keep the training and evaluation setting (\textit{i.e.}, backbone model, training corpus, evaluation benchmarks, batch size, generation budget, and decoding configurations) strictly identical to that of GRPO and UP-GRPO. Only the algorithm-specific hyperparameters in the policy-loss objective are adopted from the original papers.
\begin{table}[htbp]
\centering
\caption{Training and algorithmic hyperparameters for GRPO and UP-GRPO in the comprehensive baseline comparison on Qwen3-8B trained on MATH (level 3-5).}
\label{tab:baseline_training_config}
\begin{tabular}{lcc}
\toprule
\textbf{Item} & \textbf{GRPO} & \textbf{UP-GRPO} \\
\midrule
Prompt / Response max & 1,024 / 3,072 & 1,024 / 3,072 \\
Global batch size (prompts) & 128 & 128 \\
Rollout $n$ & 8 & 8 \\
PPO mini-batch size & 32 & 32 \\
Learning rate & $1 \times 10^{-6}$ & $1 \times 10^{-6}$ \\
Lower Clip $\epsilon_{\text{low}}$ & 0.2 & 0.2 \\
Upper Clip $\epsilon_{\text{high}}$ & 0.2 & $-$ \\
\bottomrule
\end{tabular}
\end{table}

\begin{table}[htbp]
\centering
\caption{Decoding and rollout hyperparameters for GRPO and UP-GRPO in the comprehensive baseline comparison on Qwen3-8B trained on MATH (level 3-5).}
\label{tab:baseline_decoding_config}
\begin{tabular}{lc}
\toprule
\textbf{Item} & \textbf{GRPO / UP-GRPO} \\
\midrule
Training rollout temperature & 1.0 \\
Training rollout Top-$p$ & 1.0 \\
Validation Top-$p$ & 0.7 \\
Top-$k$ & $-1$ \\
Max generation tokens & 3,072 \\
\bottomrule
\end{tabular}
\end{table}

\textbf{Table\,\ref{tab:vlm_training_config}} and \textbf{Table\,\ref{tab:vlm_decoding_config}} provide the training and decoding configurations for the multimodal experiment using Qwen3-VL-8B-Instruct on Geometry3K. Both GRPO and UP-GRPO share identical configurations; the only difference is that UP-GRPO removes the upper clipping bound for positive advantages, eliminating the need for the upper clip hyperparameter $\epsilon_{\text{high}}$.

\begin{table}[htbp]
\centering
\caption{Training and algorithmic hyperparameters for multimodal reasoning (GRPO and UP-GRPO on Qwen3-VL-8B-Instruct trained on Geometry3K).} 
\label{tab:vlm_training_config}
\begin{tabular}{lcc}
\toprule
\textbf{Item} & \textbf{GRPO} & \textbf{UP-GRPO} \\
\midrule
Prompt / Response max & 1,024 / 2,048 & 1,024 / 2,048 \\
Global batch size (prompts) & 512 & 512 \\
Rollout $n$ & 5 & 5 \\
PPO mini-batch size & 128 & 128 \\
Learning rate & $1 \times 10^{-6}$ & $1 \times 10^{-6}$ \\
Lower Clip $\epsilon_{\text{low}}$ & 0.2 & 0.2 \\
Upper Clip $\epsilon_{\text{high}}$ & 0.28 & $-$ \\
\bottomrule
\end{tabular}
\end{table}

\begin{table}[htbp]
\centering
\caption{Decoding and rollout hyperparameters for multimodal training (GRPO and UP-GRPO on Qwen3-VL-8B-Instruct trained on Geometry3K).}
\label{tab:vlm_decoding_config}
\begin{tabular}{lc}
\toprule
\textbf{Item} & \textbf{GRPO / UP-GRPO} \\
\midrule
Training rollout temperature & 1.0 \\
Training rollout Top-$p$ & 1.0 \\
Validation Top-$p$ & 0.7 \\
Top-$k$ & $-1$ \\
Max generation tokens & 2,048 \\
\bottomrule
\end{tabular}
\end{table}
\clearpage
\section{Derivation of Probability Capacity for Negative Advantages}
\label{app:neg_cap}

In this section, we derive the Probability Capacity (\text{Cap}) for the negative advantage regime ($\hat{A} \le 0$) as analyzed in Sec.\,\ref{sec:method} and visualized in Fig.\,\ref{fig:prob_capacity_comparison}(c).

For tokens with non-positive advantages, both DAPO and UP-DAPO employ a decoupled clipping mechanism to prevent excessive penalization and training instability. The clipped objective $\mathcal{J}_{\text{neg}}$ is defined as:
\begin{equation}
\mathcal{J}_{\text{neg}}(\theta) = \min \left( r_{i,t}(\theta) \hat{A}, \text{clip}(r_{i,t}(\theta), 1 - \epsilon_{\text{low}}, 1 + \epsilon_{\text{high}}) \hat{A} \right)
\end{equation}
An additional Dual Clip constraint $c$ is often introduced to bound the ratio from above when the advantage is negative. Combining these constraints, the gradient $\nabla_\theta \mathcal{J}_{\text{neg}}$ remains non-zero only when the probability ratio $r_{i,t}(\theta)$ stays within the effective optimization window:
\begin{equation}
(1 - \epsilon_{\text{low}}) \le r_{i,t}(\theta) \le c \label{eq:neg_range}
\end{equation}
where $1 - \epsilon_{\text{low}}$ is the lower trust-region boundary and $c$ is the dual-clip threshold.

To derive the Capacity, we map the constraints in Eq.\,\ref{eq:neg_range} to the absolute probability $\pi_\theta$. By multiplying the historical policy $\pi_{\text{old}}$, we identify two critical boundaries:
\begin{enumerate}
    \item \textbf{Lower Clip:} $\pi_{\text{lower}} = (1 - \epsilon_{\text{low}}) \pi_{\text{old}}$. If $\pi_\theta < \pi_{\text{lower}}$, the gradient is nullified to prevent the probability from dropping too low (over-penalization).
    \item \textbf{Dual Clip:} $\pi_{\text{upper}} = c \cdot \pi_{\text{old}}$. If $\pi_\theta > \pi_{\text{upper}}$, the gradient vanishes to prevent the policy from moving further away from the reference when the action is already deemed ``wrong.''
\end{enumerate}

In the negative regime, the goal of optimization is to \textit{decrease} the probability of the token. Thus, the Capacity represents the maximum allowable decrease before the policy hits the lower clip boundary. 

Based on the boundaries derived above, the Probability Capacity for $\hat{A} \le 0$ is formulated as the following piecewise function:
\begin{equation}
\text{Cap}(\pi_\theta, \pi_{\text{old}}) = \begin{cases} 
\pi_\theta - (1 - \epsilon_{\text{low}}) \pi_{\text{old}} & \text{if } (1 - \epsilon_{\text{low}}) \pi_{\text{old}} \le \pi_\theta \le c \cdot \pi_{\text{old}} \\ 
0 & \text{if } \pi_\theta < (1 - \epsilon_{\text{low}}) \pi_{\text{old}} \quad \text{(Lower Clip)} \\
0 & \text{if } \pi_\theta > c \cdot \pi_{\text{old}} \quad \text{(Dual Clip)}
\end{cases}
\end{equation}

This derivation explains the blue regions in Fig.\,\ref{fig:prob_capacity_comparison}(c). The Cap is maximized when $\pi_\theta$ is near the dual-clip boundary and gradually diminishes to zero as $\pi_\theta$ approaches the trust-region lower limit.
\clearpage
\section{Derivation of the UP-GSPO Gradient}
\label{app:up_gspo_derivation}

To rigorously establish the gradient behavior of UP-GSPO for positive advantage samples ($\hat{A}_i > 0$), we derive the exact analytical gradient of its sequence-level surrogate objective. 

For a given prompt $q$ and a set of $G$ sampled responses $\{o_i\}_{i=1}^G$, we define the self-anchored sequence-level Unbounded Positive (UP) objective as follows:
\begin{equation}
    \mathcal{J}_{\text{UP-GSPO}}^+(\theta) = \mathbb{E}_{q \sim \mathcal{Q}, \{o_i\}_{i=1}^G \sim \pi_{\text{old}}} \left[ \frac{1}{G} \sum_{i=1}^G \hat{A}_i \left( \frac{\pi_\theta(o_i|q)}{\text{sg}(\pi_\theta(o_i|q))} \right)^{\frac{1}{|o_i|}} \right]
\end{equation}
where $\text{sg}(\cdot)$ denotes the stop-gradient operator, which strictly treats its operand as a constant during backpropagation.

When applying the gradient operator $\nabla_\theta$, the denominator $\text{sg}(\pi_\theta(o_i|q))$ factors out as a constant. Because the forward numerical value of the stop-gradient term is strictly identical to the policy probability (i.e., $\text{sg}(\pi_\theta) \equiv \pi_\theta$), we can seamlessly substitute this equivalence back into the expression after applying the chain rule $\nabla_\theta(f^\alpha) = \alpha f^{\alpha-1} \nabla_\theta f$. This yields a continuous and elegant derivation:
\begin{align} 
    \nabla_\theta \mathcal{J}_{\text{UP-GSPO}}^+(\theta) 
    &= \mathbb{E}_{q, \{o_i\}} \left[ \frac{1}{G} \sum_{i=1}^G \hat{A}_i \frac{1}{\text{sg}(\pi_\theta(o_i|q))^{\frac{1}{|o_i|}}} \nabla_\theta \left( \pi_\theta(o_i|q)^{\frac{1}{|o_i|}} \right) \right] \nonumber \\
    &= \mathbb{E}_{q, \{o_i\}} \left[ \frac{1}{G} \sum_{i=1}^G \hat{A}_i \frac{1}{\pi_\theta(o_i|q)^{\frac{1}{|o_i|}}} \left( \frac{1}{|o_i|} \pi_\theta(o_i|q)^{\frac{1}{|o_i|} - 1} \nabla_\theta \pi_\theta(o_i|q) \right) \right] \nonumber \\
    &= \mathbb{E}_{q, \{o_i\}} \left[ \frac{1}{G} \sum_{i=1}^G \hat{A}_i \frac{1}{|o_i|} \frac{\nabla_\theta \pi_\theta(o_i|q)}{\pi_\theta(o_i|q)} \right] \label{eq:up_gspo_grad_simplified}
\end{align}
Notice how the complex length-normalization exponents algebraically cancel out perfectly ($-\frac{1}{|o_i|} + \frac{1}{|o_i|} - 1 = -1$). 

Finally, by applying the log-derivative trick ($\frac{\nabla_\theta \pi}{\pi} = \nabla_\theta \log \pi$) and expanding the sequence-level joint probability into the sum of its token-level log probabilities ($\log \pi_\theta(o_i|q) = \sum_{t=1}^{|o_i|} \log \pi_\theta(o_{i,t}|q, o_{i,<t})$), the expected gradient simplifies to our final form:
\begin{equation} \label{eq:up_gspo_grad_final}
    \nabla_\theta \mathcal{J}_{\text{UP-GSPO}}^+(\theta) = \mathbb{E}_{q \sim \mathcal{Q}, \{o_i\}_{i=1}^G \sim \pi_{\text{old}}} \left[ \frac{1}{G} \sum_{i=1}^G \hat{A}_i \left( \frac{1}{|o_i|} \sum_{t=1}^{|o_i|} \nabla_\theta \log \pi_\theta(o_{i,t}|q, o_{i,<t}) \right) \right]
\end{equation}

\textbf{Remark:} Equation~\ref{eq:up_gspo_grad_final} rigorously demonstrates that for positive advantages, the UP-GSPO sequence-level objective mathematically equates to a length-normalized REINFORCE gradient. By completely bypassing the dynamic importance sampling ratio, this formulation explicitly removes the clipping upper bound, maximizing exploration capacity while cleanly preserving the variance-reducing properties of the $\frac{1}{|o_i|}$ normalization.

Building upon this derivation, we formulate the final, unified UP-GSPO objective. Recognizing the divergent optimization dynamics between correct and incorrect rollouts, UP-GSPO employs an asymmetric routing mechanism. We retain the standard sequence-level clipped objective for negative advantages to act as a structural safeguard, while deploying our Unbounded Positive objective for positive advantages. 

Let $s_i(\theta) = \left( \frac{\pi_\theta(o_i|q)}{\pi_{\text{old}}(o_i|q)} \right)^{\frac{1}{|o_i|}}$ denote the length-normalized sequence-level importance weight. The final UP-GSPO objective is formulated as:
\begin{equation} \label{eq:final_up_gspo}
    \mathcal{J}_{\text{UP-GSPO}}(\theta) = \mathbb{E}_{q \sim \mathcal{Q}, \{o_i\} \sim \pi_{\text{old}}} \left[ \frac{1}{G} \sum_{i=1}^G \begin{cases} 
        \hat{A}_i \left( \frac{1}{|o_i|} \sum_{t=1}^{|o_i|} \log \pi_\theta(o_{i,t}|q, o_{i,<t}) \right) & \text{if } \hat{A}_i > 0 \\
        \min \Big( s_i(\theta) \hat{A}_i, \, \text{clip}(s_i(\theta), 1 - \epsilon, 1 + \epsilon) \hat{A}_i \Big) & \text{if } \hat{A}_i \le 0
    \end{cases} \right]
\end{equation}

\end{document}